\documentclass[10pt,twocolumn,letterpaper]{article}

\usepackage[pagenumbers]{cvpr} 

\usepackage{graphicx}
\usepackage{amsmath}
\usepackage{amssymb}
\usepackage{booktabs}

\usepackage{microtype}
\usepackage{xcolor}
\newcommand{\oo}[1]{\textcolor{orange}{#1}}

\usepackage{amsmath,amsfonts,bm}

\def\eqref#1{equation~\ref{#1}}

\def\1{\bm{1}}

\def\va{{\bm{a}}}

\def\vn{{\bm{n}}}

\def\vp{{\bm{p}}}

\def\vr{{\bm{r}}}

\def\vx{{\bm{x}}}

\def\mA{{\bm{A}}}

\def\mN{{\bm{N}}}

\def\mP{{\bm{P}}}

\def\mX{{\bm{X}}}

\DeclareMathAlphabet{\mathsfit}{\encodingdefault}{\sfdefault}{m}{sl}
\SetMathAlphabet{\mathsfit}{bold}{\encodingdefault}{\sfdefault}{bx}{n}

\DeclareMathOperator*{\argmax}{arg\,max}
\DeclareMathOperator*{\argmin}{arg\,min}

\DeclareMathOperator{\sign}{sign}

\usepackage{pifont}

\usepackage{enumitem}
\usepackage{multirow}

\usepackage{slashbox}
\usepackage{booktabs}
\usepackage{colortbl}
\usepackage{soul}
\usepackage{comment}

\definecolor{crimson}{HTML}{DC143C}
\definecolor{tab:gray}{HTML}{7f7f7f}
\definecolor{tab:pink}{HTML}{e377c2}
\definecolor{tab:cyan}{HTML}{17becf}
\definecolor{tab:orange}{HTML}{ff7f0e}
\definecolor{tab:blue}{HTML}{1f77b4}
\definecolor{tab:red}{HTML}{d62728}

\usepackage[pagebackref,breaklinks,colorlinks]{hyperref}

\usepackage[capitalize]{cleveref}
\crefname{section}{Sec.}{Secs.}
\Crefname{section}{Section}{Sections}
\Crefname{table}{Table}{Tables}
\crefname{table}{Tab.}{Tabs.}

\def\cvprPaperID{201} 
\def\confName{CVPR}
\def\confYear{2022}

\begin{document}

\title{Enhancing Adversarial Robustness for Deep Metric Learning}

\author{Mo Zhou\\
Johns Hopkins University\\
{\tt\small mzhou32@jhu.edu}
\and
Vishal M. Patel\\
Johns Hopkins University\\
{\tt\small vpatel36@jhu.edu}
}
\maketitle

\begin{abstract}
	Owing to security implications of adversarial vulnerability, adversarial
	robustness of deep metric learning models has to be improved.
	In order to avoid model collapse due to excessively hard examples, the
	existing defenses dismiss the min-max adversarial training, but instead
	learn from a weak adversary inefficiently.
	Conversely, we propose Hardness Manipulation to efficiently perturb the
	training triplet till a specified level of hardness for adversarial
	training, according to a harder benign triplet or a pseudo-hardness
	function.
	It is flexible since regular training and min-max adversarial training
	are its boundary cases.
	Besides, Gradual Adversary, a family of pseudo-hardness functions is
	proposed to gradually increase the specified hardness level during training
	for a better balance between performance and robustness.
	Additionally, an Intra-Class Structure loss term among benign and adversarial
	examples further improves model robustness and efficiency.
	Comprehensive experimental results suggest that the proposed method,
	although simple in its form, overwhelmingly outperforms the
	state-of-the-art defenses in terms of robustness, training efficiency,
	as well as performance on benign examples.
\end{abstract}

\section{Introduction}
\label{sec:1}


Given a set of data points, a \emph{metric} gives a distance value between each
pair of them.
Deep Metric Learning (DML) aims to learn such a metric between two inputs (\eg,
images) leveraging the representational power of deep neural networks.
As an extensively studied task~\cite{revisiting,dmlreality}, DML has a wide
range of applications such as image retrieval~\cite{imagesim2} and face
recognition~\cite{facenet,domainface}, and widely influences some other areas
such as self-supervised learning~\cite{dmlreality}.

Despite the advancements in this field thanks to deep learning, recent studies
find DML models vulnerable to adversarial attacks, where  imperceptible
perturbations can incur unexpected retrieval result, or covertly change the
rankings~\cite{advrank,advorder}.
Such vulnerability raises security, safety, and fairness concerns in the DML
applications.
For example, impersonation or recognition evasion are possible on a vulnerable
DML-based face-identification system.
To counter the attacks (\ie, mitigating the vulnerability), the
\emph{adversarial robustness} of DML models has to be improved via defense.

\begin{figure}[t]
	\includegraphics[width=1.0\columnwidth]{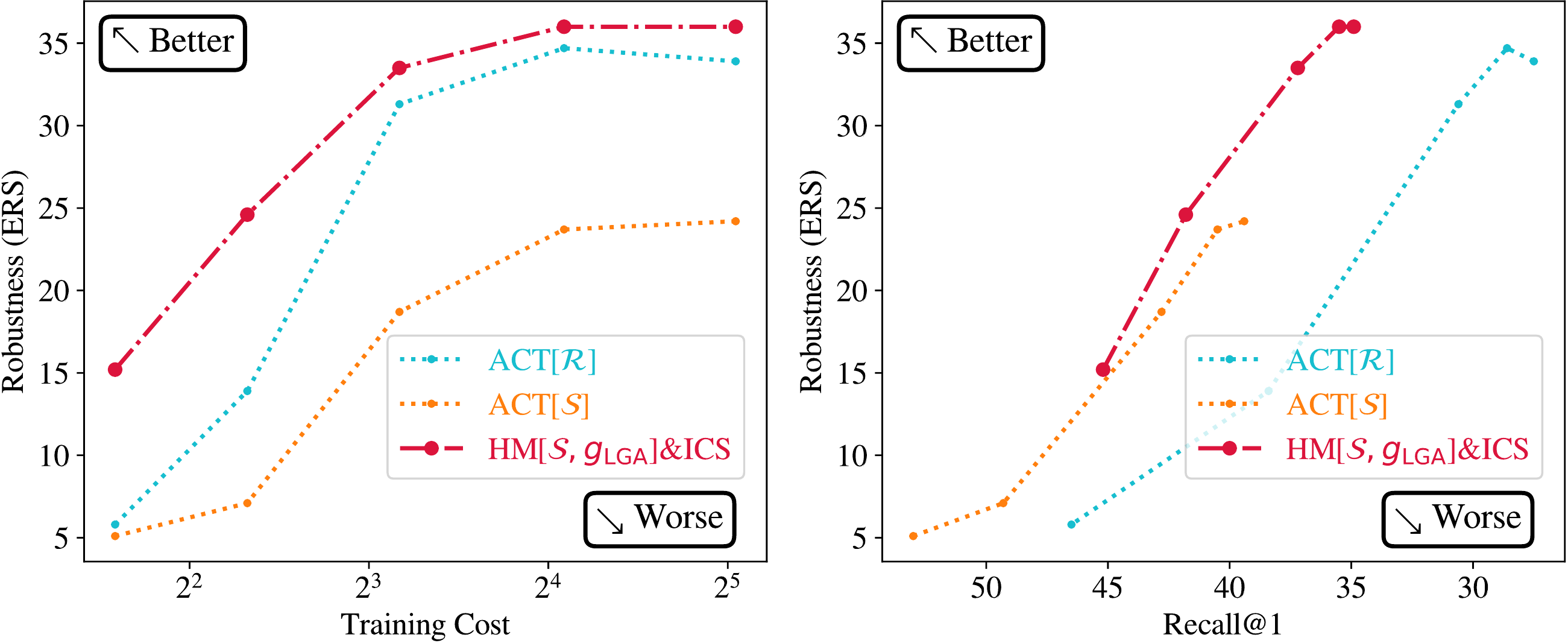}
	\caption{
		Comparison in robustness, training cost, and recall@1 
		between our method (\ie, ``HM[$\mathcal{S},g_\mathsf{LGA}$]\&ICS'')
		and the state-of-the-art method (\ie, ``ACT[$\mathcal{R}$]'' and
		``ACT[$\mathcal{S}$]'') on the CUB Dataset.
	}
	\label{fig:introplot}
\end{figure}


Existing defense methods~\cite{advrank,robrank} are adversarial training-based,
inspired by Madry's \emph{min-max} adversarial training~\cite{madry} because it
is consistently one of the most effective methods for classification task.
Specifically, Madry's method involves a inner problem to \emph{maximize} the
loss by perturbing the inputs into adversarial examples, and an outer problem
to \emph{minimize} the loss by updating the model parameters.
However, in order to avoid model collapse due to excessively hard examples, the
existing DML defenses refrain from directly adopting such min-max paradigm, but
instead replace the inner problem to indirectly increase the loss value to a
certain level, which suffers from low efficiency and weak adversary (and hence
weak robustness).
Since training cost is already a serious issue of adversarial training, the
efficiency in gaining higher adversarial robustness under a lower budget is
inevitable and important for DML defense.


Inspired by previous works~\cite{advrank,robrank}, we conjecture that an
appropriate adversary for the inner \emph{maximization} problem should
increase the loss to an ``intermediate''
point between that of benign examples (\ie, unperturbed examples) and the
theoretical upper bound.
Such point should be reached by an efficient adversary directly.
Besides, we speculate the triplet sampling strategy has a key impact in
adversarial training, because it is also able to greatly influence the
mathematical expectation of loss even without adversarial attack.

\begin{figure}
	\includegraphics[width=\columnwidth]{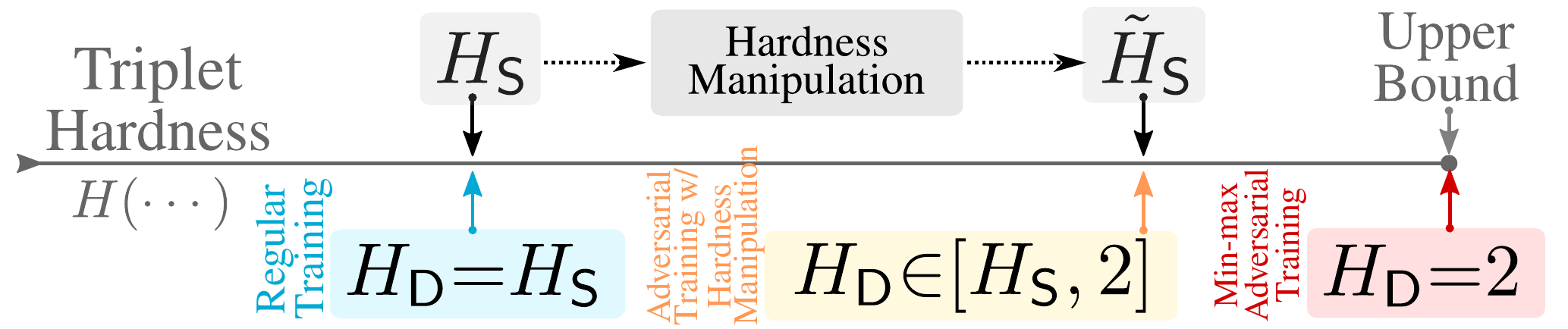}
	\vspace{-1.8em}
	\caption{Flexibility of hardness manipulation.
	Regular training and min-max adversarial training are its boundary cases.
	}
	\label{fig:hmflexible}
\end{figure}

In this paper, we first define the ``\emph{hardness}'' of a sample triplet as
the difference between the anchor-positive distance and anchor-negative
distance.
Then, Hardness Manipulation (HM) is proposed to adversarially perturb a given
sample triplet and increase its hardness into a specified \emph{destination}
hardness level for adversarial training.
The objective of HM is to minimize the L-$2$ norm of the thresholded difference
between the hardness of the given sample triplet and the specified
\emph{destination} hardness.
HM is flexible as regular training and min-max adversarial training~\cite{madry}
can be expressed as its boundary cases, as shown in \cref{fig:hmflexible}.
Mathematically, when the HM objective is optimized using Projected Gradient
Descent~\cite{madry}, the sign of its gradient with respect to the adversarial
perturbation is the same as that of directly \emph{maximizing} the loss.
Thus, the optimization of HM objective can be interpreted as a direct and
efficient \emph{maximization} process of the loss which stops halfway at the
specified \emph{destination} hardness level, \ie, the aforementioned ``intermediate'' point.


Then, how hard should such ``\emph{destination} hardness'' be?
Recall that the model is already prone to collapse with excessively hard benign
triplets~\cite{facenet}, let alone adversarial examples.
Thus, intuitively, the \emph{destination} hardness can be the hardness of
another benign triplet which is moderately harder than the given triplet (\eg,
a Semihard~\cite{facenet} triplet).
However, in the late phase of training, the expectation of the difference
between such \emph{destination} hardness and that of the given triplet will be
small, leading to weak adversarial examples and inefficient adversarial
learning.
Besides, strong adversarial examples in the early phase of training may also
hinder the model from learning good embeddings, and hence influence the
performance on benign examples.
In particular, a better \emph{destination} hardness should be able to balance the
training objectives in the early and late phases of training.


To this end, Gradual Adversary, a family of pseudo-hardness functions is
proposed, which can be used as the \emph{destination} hardness.
A function that leads to relatively weak and relatively strong adversarial
examples, respectively in the early and late phase of training belongs to this
family.
As an example, we design a ``Linear Gradual Adversary'' (LGA)
function as the linearly scaled negative triplet margin, incorporating a strong
prior that the \emph{destination} hardness should remain Semihard
based on our empirical observation.


Additionally, it is noted that a sample triplet will be augmented into a
sextuplet (both benign and adversarial examples) during adversarial training.
In this case, the \emph{intra-class} structure can be enforced, which has been
neglected by existing methods.
Since some existing attacks aim to change the sample rankings in the same
class~\cite{advrank},
we propose a simple \emph{intra-class} structure loss term for adversarial
training, which is expected to further improve adversarial robustness.

Comprehensive experiments are conducted on three commonly used DML datasets,
namely CUB-200-2011~\cite{cub200}, Cars-196~\cite{cars196}, and Stanford Online
Product~\cite{sop}.
The proposed method overwhelmingly outperforms the state-of-the-art defense in
terms of robustness, training efficiency, as well as the performance on benign
examples.

In summary, our contributions include proposing:
\begin{enumerate}[nosep, noitemsep, leftmargin=*]
	\item {\textit{Hardness Manipulation}} (HM) as a flexible and efficient
		tool to create adversarial example triplets for subsequent adversarial
		training of a DML model.
	\item \textit{Linear Gradual Adversary} (LGA) as a Gradual Adversary, \ie,
		a pseudo-hardness function for HM, which incorporates our empirical
		observations and can balance the training objectives during the
		training process.
	\item \textit{Intra-Class Structure} (ICS) loss term to further improve
		model robustness and adversarial training efficiency, while such structure is
		neglected by existing defenses. 
\end{enumerate}

\section{Related Works}
\label{sec:2}

\textbf{Adversarial Attack.}
Szegedy \etal~\cite{l-bfgs} find misclassification of DNN can be triggered by
an imperceptible adversarial perturbation to the input image.
Ian \etal~\cite{fgsm} attribute the reason to DNN being locally linear with
respect to the adversarial perturbation.
Subsequent first-order gradient-based methods can compromise the DNNs more
effectively under the white-box assumption~\cite{i-fgsm,madry,apgd,lafeat}.
In contrast, black-box attacks have been explored by query-based
methods~\cite{nes-atk,spsa-atk} and transferability-based
methods~\cite{di-fgsm}, which are more practical for real-world scenarios.

\textbf{Adversarial Defense.}
Various defenses are proposed to counter the attacks.
However, defenses incurring gradient masking
lead to a false sense of robustness~\cite{obfuscated}.
Defensive distillation~\cite{distill2} is compromised in~\cite{cw}.
Ensemble of weak defenses is not
robust~\cite{ensembleweak}.
Other defenses such as input
preprocessing~\cite{deflecting}, or randomization~\cite{self-ensemble} are proposed.
But many of them are still susceptible to adaptive attacks~\cite{adaptive}.
Of all defenses, adversarial training~\cite{madry} consistently remains to be one of the
most effective
methods~\cite{bilateral,advtrain-triplet,benchmarking,smoothat,trades,robustwrn,weightperturb,featurescatter,currat,awp},
but suffers from high training cost~\cite{freeat,fastat,yopo}, performance drop on
benign examples~\cite{odds,geometry,onmanifold}, and overfitting on adversarial examples~\cite{bagoftricks,overfitting}.

\textbf{Deep Metric Learning.}
A wide range of applications such as image retrieval~\cite{imagesim2},
cross-modal retrieval~\cite{ladderloss}, and face recognition~\cite{facenet}
can be formulated as a DML problem.
A well-designed loss function and a proper sampling method are crucial for DML
performance~\cite{dmlreality}.  For instance, the classical
triplet loss~\cite{facenet} could reach state-of-the-art performance with an appropriate
sampling strategy~\cite{revisiting}.

\textbf{Attacks in DML.}
DML has been found vulnerable to adversarial attacks as
well~\cite{advrank,advorder,robrank}, which raises concerns on safety,
security, or fairness for a DML application.
The existing attacks aim to completely subvert the image retrieval
results~\cite{qair,learn-to-misrank,advdpqn,advpattern,flowertower,universalret},
or covertly alter the top-ranking results without being
abnormal~\cite{advrank,advorder}.

\textbf{Defenses in DML.} Unlike attacks, defenses are less explored.
Embedding Shifted Triplet (EST)~\cite{advrank} is an
adversarial training method using adversarial examples with maximized embedding
move distance off their original locations.
The state-of-the-art method, \ie,
Anti-Collapse Triplet (ACT)~\cite{robrank} forces the model to separate
collapsed positive and negative samples apart in order to learn robust
features.
However, both EST and ACT suffer from low efficiency as the
inner problem is replaced into an indirect adversary.

\section{Our Approach}
\label{sec:3}


In DML~\cite{revisiting,dmlreality}, a 
function $\phi:\mathcal{X}{\mapsto}\Phi \subseteq \mathbb{R}^D$ is learned to
map data points $\mX\in\mathcal{X}$ into an embedding space $\Phi$, which is usually
normalized to the real unit hypersphere for regularization.
With a predefined distance function $d(\cdot,\cdot)$, which is usually the
Euclidean distance, we can measure the distance between $\mX_i$ and $\mX_j$ as
$d_\phi(\mX_i,\mX_j)=d(\phi(\mX_i),\phi(\mX_j))$.
Typically, the triplet loss~\cite{facenet} can be used to learn the
embedding function, and it could reach the state-of-the-art performance with an
appropriate triplet sampling strategy~\cite{revisiting}.


Given a triplet of anchor, positive, negative images, \ie, $\mA, \mP,
\mN \in \mathcal{X}$, we can calculate their embeddings with $\phi(\cdot)$ as
$\va, \vp, \vn$, respectively.
Then triplet loss~\cite{facenet} is defined as:
\begin{equation}
	L_\text{T}(\va, \vp, \vn; \gamma) = \max(0, d(\va, \vp) - d(\va, \vn) +
	\gamma),
	\end{equation}
%
%
where $\gamma$ is a predefined margin parameter.
To attack the DML model, an imperceptible adversarial perturbation $\vr \in
\Gamma$ is added to the input image $\mX$, where $\Gamma = \{\vr | \mX+\vr\in
\mathcal{X},  \|\vr\|_p \leq \varepsilon\}$, so that its embedding vector
$\tilde{\vx}=\phi(\mX+\vr)$ will be moved off its original location towards
other positions to achieve the attacker's goal.
To defend against the attacks, the DML model can be adversarially trained to
reduce the effect of attacks~\cite{advrank,robrank}.
The most important metrics for a good defense are adversarial
robustness, training efficiency, and performance on benign examples.

\subsection{Hardness Manipulation}
\label{sec:31}


Given an image triplet ($\mA$, $\mP$, $\mN$) sampled with a certain sampling
strategy (\eg, Random) within a mini-batch, we define its ``\emph{hardness}''
as a scalar which is within $[-2,2]$:
\begin{equation}
H(\mA,\mP,\mN)=d_\phi(\mA,\mP)-d_\phi(\mA,\mN).
\end{equation}
Clearly, it is an internal part of the triplet loss.
For convenience, we call this triplet ($\mA$, $\mP$, $\mN$) as ``\emph{source}
triplet'', and its corresponding hardness value as ``\emph{source} hardness'',
denoted as $H_\mathsf{S}$.

\begin{figure}
	\includegraphics[width=\columnwidth]{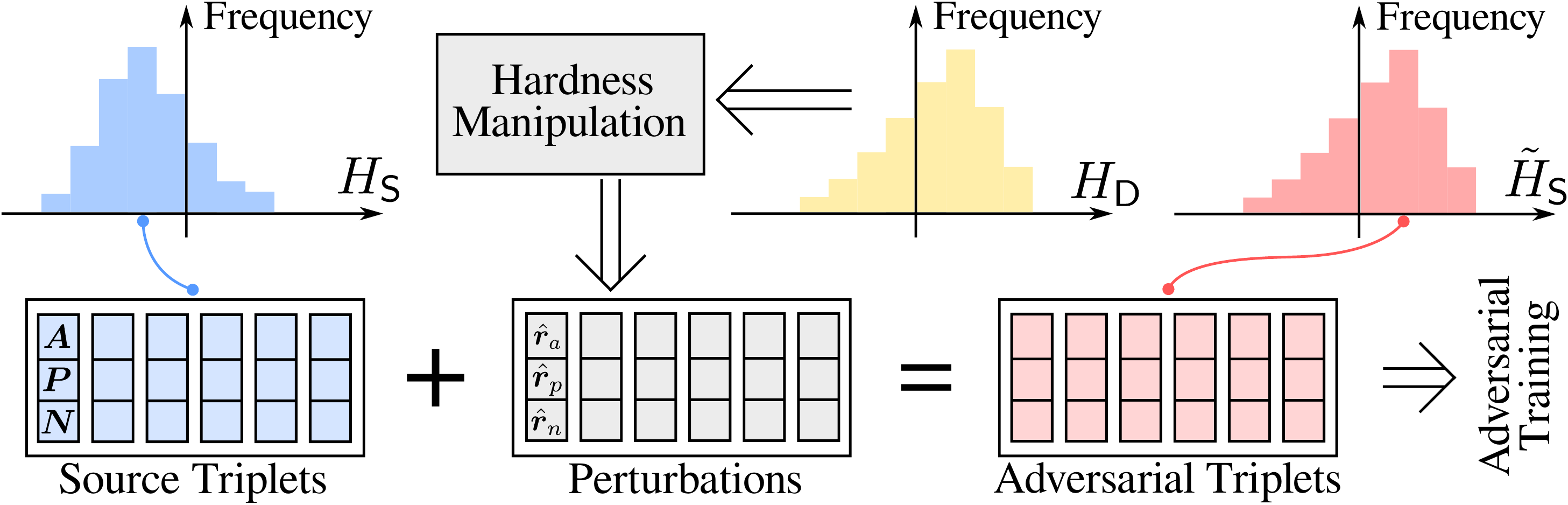}
	\vspace{-1.8em}
	\caption{Illustration of hardness manipulation.}
	\label{fig:hm}
\end{figure}

Then, \emph{Hardness Manipulation} (HM) aims to increase the \emph{source}
hardness $H_\mathsf{S}$ into a specified ``\emph{destination} hardness''
$H_\mathsf{D}$, by finding adversarial examples of the source triplet, \ie,
$(\mA{+} {\vr}_a, \mP {+} {\vr}_p, \mN {+} {\vr}_n)$, where $({\vr}_a, {\vr}_p,
{\vr}_n)$ are the adversarial perturbations.
Denoting the hardness of the adversarially perturbed \emph{source} triplet
as $\tilde{H}_\mathsf{S}$, \ie,
$\tilde{H}_\mathsf{S} = H(\mA{+} {\vr}_a, \mP {+} {\vr}_p, \mN {+} {\vr}_n)$,
the HM is implemented as:
\begin{equation}
	\hat{\vr}_a, \hat{\vr}_p, \hat{\vr}_n = \argmin_{\vr_a,\vr_p,\vr_n}
	\big\|\max(0, H_\mathsf{D} - \tilde{H}_\mathsf{S} ) \big\|_2^2.
	\label{eq:hm}
\end{equation}
The $\max(0,\cdot)$ part in \cref{eq:hm} truncates the gradient when
$\tilde{H}_\mathsf{S}>H_\mathsf{D}$, automatically stopping the optimization, because
$\tilde{H}_\mathsf{S}$ is not desired to be reduced once it exceeds $H_\mathsf{D}$.
 \cref{eq:hm} is written in the L-$2$ norm form instead of the standard Mean
Squared Error because HM can be directly extended into vector form for a
mini-batch.
The optimization problem can be solved by Projected Gradient Descent
(PGD)~\cite{madry}.
And the resulting adversarial examples are used for adversarially training the
DML model with
$L_\text{T}(\phi(\mA+\hat{\vr}_a), \phi(\mP+\hat{\vr}_p),
\phi(\mN+\hat{\vr}_n))$.
The overall procedure of HM is illustrated in \cref{fig:hm}.
For convenience,
we abbreviate the adversarial training with adversarial examples created
through this way as ``$\text{HM}[H_\mathsf{S},H_\mathsf{D}]$'' in this paper.

Note, in the PGD case, the sign of negative gradient of the HM objective
\emph{w.r.t.} an adversarial perturbation $\vr$ is equivalent to the sign of
gradient for directly maximizing $\tilde{H}_\mathsf{S}$ (hence
maximizing $L_\text{T}$) when $H_\mathsf{D}>\tilde{H}_\mathsf{S}$, \ie,
\begin{align}
	\Delta \vr
	=&
	\sign\big\{
		-\frac{\partial}{\partial \vr} \big\| \max(0, 
		H_\mathsf{D} - \tilde{H}_\mathsf{S} ) \big\|_2^2
	\big\}
	\\
	=&
	\sign\big\{
		2(H_\mathsf{D} - \tilde{H}_\mathsf{S})
		\frac{\partial }{\partial \vr} \tilde{H}_\mathsf{S} \big\}
	=
	\sign\big\{
		\frac{\partial}{\partial \vr} \tilde{H}_\mathsf{S}
		\big\}.
\end{align}
The perturbation $\vr$ is updated as $\vr \leftarrow \text{Proj}_\Gamma\{\vr + \alpha \Delta
\vr\}$ by PGD for $\eta$ steps with a step size $\alpha$, where the ``Proj'' operator
clips the result into the $\Gamma$ set.
Thus, the optimization of HM objective can be interpreted as direct
maximization of $\tilde{H}_\mathsf{S}$, which discontinues very early once it
exceeds $H_\mathsf{D}$.
With HM, the model can learn from an \emph{efficient} adversary.

Since the same $\Delta\vr$ can be used for both minimizing the HM objective
and maximizing the triplet loss, one potential advantage of HM is that
the gradients during the training process can be reused for creating
adversarial examples for much faster adversarial training, according to 
Free Adversarial Training~\cite{freeat}.
We leave this for future exploration.


\textbf{Destination Hardness}.
$\text{HM}[H_\mathsf{S},H_\mathsf{D}]$ is flexible as various
types of $H_\mathsf{D}$ can be specified, \eg, a constant, the
hardness of another benign triplet, or a pseudo-hardness function.
The case of maximizing the triplet loss is equivalent to
$\text{HM}[H_\mathsf{S},2]$, where $2$ is the 
upper bound of hardness, while $\text{HM}[H_\mathsf{S},H_\mathsf{S}]$
is regular DML training, as shown in \cref{fig:hmflexible}.

As pushing $\tilde{H}_\mathsf{S}$ towards the upper bound will easily render
model collapse, a valid $H_\mathsf{D}$ should be chosen within the interval
$[H_\mathsf{S},2]$.
Thus, intuitively, $H_\mathsf{D}$ can be the hardness of another benign triplet
(with the same anchor) sampled with a strategy with a higher hardness
expectation, \ie, $E[H_\mathsf{D}] > E[H_\mathsf{S}]$.
Or at least the $\text{Var}[H_\mathsf{D}]$ of another benign triplet has to
be large enough (for a small portion of triplets $H_\mathsf{D}>H_\mathsf{S}$)
in order to create a notable number of valid adversarial
examples.
For instance, $H_\mathsf{D}$ can be the hardness of a Semihard~\cite{facenet}
triplet when the \emph{source} triplet is sampled with Random sampler.
Predictably, the model performance will be significantly influenced by the
triplet sampling strategies we chose for $H_\mathsf{D}$ in this case.
For convenience of further discussion, we denote the hardness of a Random,
Semihard, and Softhard triplets as $\mathcal{R}$, $\mathcal{M}$, $\mathcal{S}$,
respectively.

If we have a strong prior knowledge on what the \emph{destination} hardness
should be, then we can even use a pseudo-hardness function $g(\cdot)$, \ie, a
customized scalar function.

\subsection{Gradual Adversary}
\label{sec:32}

\begin{figure}
	\includegraphics[width=\columnwidth]{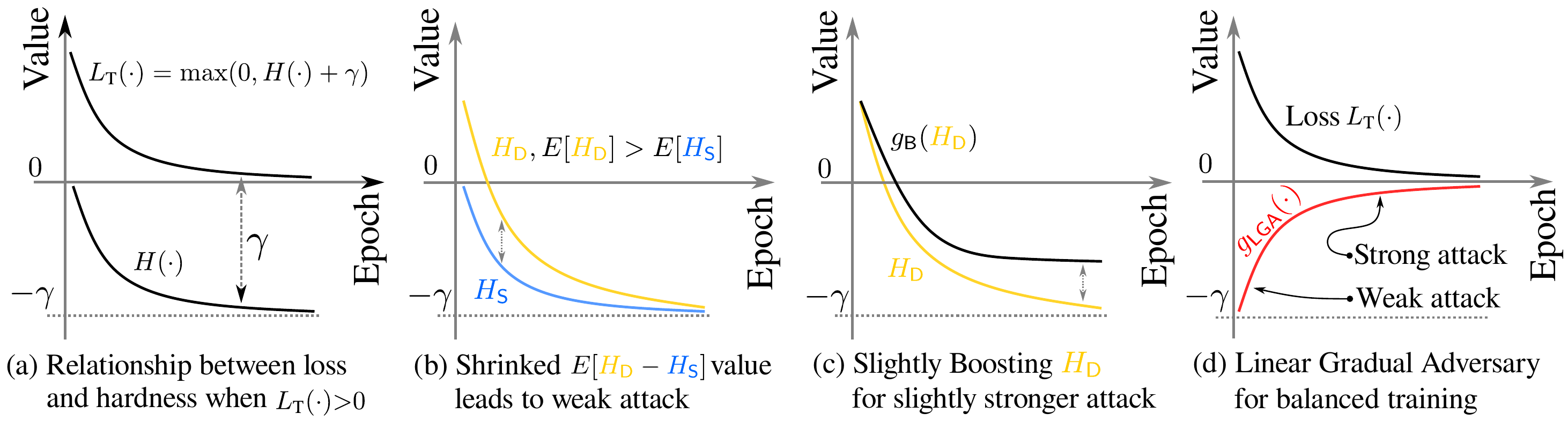}
	\caption{Illustration of (linear) gradual adversary.}
	\label{fig:ga}
\end{figure}


Even if $H_\mathsf{D}$ is calculated from another triplet harder than the
\emph{source} triplet, the adversarial example may become weak in the late
phase of training.
The optimizer aims to reduce the expectation of 
loss $E[L_\text{T}]$ towards zero as possible over the distribution of triplets, and thus
the $E[H]$ of any given triplet will tend to $-\gamma$, reducing the
hardness of adversarial triplets from HM as $E[H_\mathsf{D}-H_\mathsf{S}]$
decreases accordingly.
Weakened adversarial examples are insufficient for robustness.


Intuitively, such deficiency can be alleviated with a \emph{pseudo-hardness}
function that slightly increases the value of $H_\mathsf{D}$ in the late phase
of training.
Denoting the loss value from the previous training iteration as $\ell_{t-1}$,
we first normalize it into $[0,1]$ as $\bar{\ell}_{t-1}=\min(u,\ell_{t-1})/u$,
where $u$ is a manually specified constant.
Then we can linearly shift the $E[H_\mathsf{D}]$ by a scaled constant $\xi$,
\ie,
\begin{equation}
	g_\mathsf{B}(H_\mathsf{D};\xi,\bar{\ell}_{t-1}) =
	H_\mathsf{D} + \xi \cdot (1-\bar{\ell}_{t-1}).
\end{equation}
The deficiency can be alleviated in
$\text{HM}[H_\mathsf{S},g_\mathsf{B}(H_\mathsf{D})]$.


Apart from the deficiency in the late phase of training, we speculate that
relatively strong adversarial examples may hinder the model from 
learning good embedding space for the benign examples in the very early phase
of training, hence influence the model performance on benign examples.


Thus, $H_\mathsf{D}$ should lead to (1) relatively weak adversarial examples in
the early training phase (indicated by a large loss value), and (2) relatively
strong adversarial examples in the late training phase (indicated by a small
loss value),
in order to automatically balance the training objectives (\ie, performance on
benign examples \emph{v.s.} robustness).
A satisfactory pseudo-hardness function is a ``Gradual Adversary''.


As an example, we propose a ``Linear Gradual Adversary'' (LGA) pseudo-hardness
function that is independent to any benign triplets, incorporating our empirical
observation that $H_\mathsf{D}$ should remain Semihard~\cite{facenet}, as
follows:
\begin{equation}
	g_\mathsf{LGA}(\bar{\ell}_{t-1}) = -\gamma \cdot \bar{\ell}_{t-1} ~ \in
	[-\gamma,0].
\end{equation}
Our empirical observation is obtained from \cref{sec:41}.
And the training objectives, namely performance on benign examples and
robustness will be automatically balanced in
$\text{HM}[H_\mathsf{S},g_\mathsf{LGA}]$, leading to a better eventual overall
performance, as illustrated in \cref{fig:ga}.
More complicated or non-linear pseudo-hardness functions are left for future study.
%

\subsection{Intra-Class Structure}
\label{sec:33}

\begin{figure}[t]
	\includegraphics[width=1.0\columnwidth]{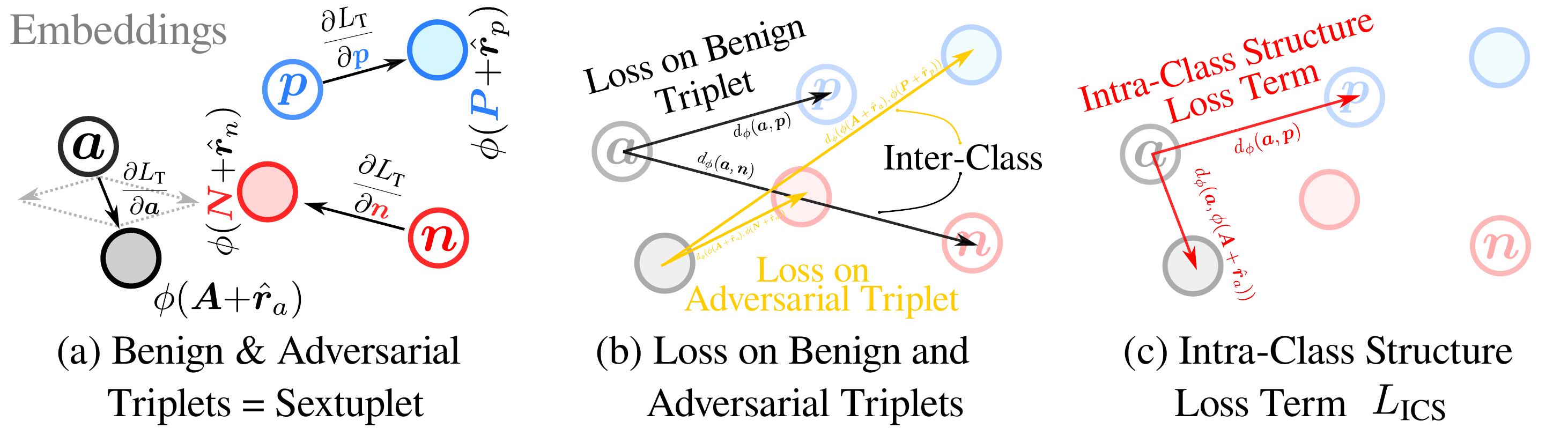}
	\caption{Illustration of intra-class structure loss term.}
	\label{fig:ics}
\end{figure}


During adversarial training with HM, the adversarial counterpart of each given sample
triplet is fed to the model, and the triplet loss will enforce a good
\emph{inter-class} structure.
Since the anchor, positive sample, and their adversarial counterpart belongs to
the same class, it should be noted that the \emph{intra-class} structure can be
enforced as well, but this has been neglected by the existing DML defenses.
\emph{Intra-class} structure is also important for robustness besides the
\emph{inter-class} structure, because the attack may attempt to change the
rankings of samples in the same class~\cite{advrank}.


We propose an additional loss function term to enforce such \emph{intra-class}
structure, as shown in \cref{fig:ics}.
Specifically, the anchor $\va$ and its adversarial counterpart
are separated from the positive sample $\vp$ by reusing the triplet loss,
\ie,
\begin{equation}
	L_\text{ICS} = \lambda \cdot L_\text{T}(
	\va, \phi(\mA + \hat{\vr}_a), \vp; 0),
\end{equation}
where $\lambda$ is a constant weight for this loss term,
and the margin is set as zero to avoid negative effect.
The $L_\text{ICS}$ term can be appended to the loss function for
adversarial training.

\section{Experiments}
\label{sec:4}


To validate our defense method, we conduct experiments
on three commonly used DML datasets: CUB-200-2011 (CUB)~\cite{cub200}, Cars-196
(CARS)~\cite{cars196}, and Stanford-Online-Product (SOP)~\cite{sop}.
We follow the same experimental setup as that used in the state-of-the-art
defense work~\cite{robrank} and standard DML~\cite{revisiting} for ease of comparison.

\begin{table}
\resizebox{\columnwidth}{!}{
	\setlength{\tabcolsep}{0.8em}
\begin{tabular}{c|ccccc}
\toprule
\rowcolor{black!7} Statistics & Random & Semihard & Softhard & Distance & Hardest\tabularnewline
\midrule
$E[H]$ & $-0.164$ & $-0.126$ & $-0.085$ & $0.043$ & $0.044$\tabularnewline
$\text{Var}[H]$ & $0.00035$ & $0.00013$ & $0.00122$ & $0.00021$ & $0.00021$\tabularnewline
\bottomrule
\end{tabular}
}
	\vspace{-0.5em}
\caption{Mean \& variance of hardness w/ various triplet samplers.
Calculated as the average statistics over $1000$ mini-batches from the CUB dataset
with an imagenet-initialized RN18 model.
}

	\label{tab:hsort}
\end{table}


Specifically, we (adversarially) train ImageNet-initialized ResNet-18
(RN18)~\cite{resnet} with the output dimension of the last layer changed to
$N{=}512$.
The margin $\gamma$ in the triplet loss is $0.2$.
Adam~\cite{adam} optimizer is used for parameter updates, with a learning rate
of $1.0{\times}10^{-3}$ for $150$ epochs, and a mini-batch size of $112$.
Adversarial examples are created within $\Gamma$ with
$\varepsilon{=}8/255$ and $p{=}\infty$, using PGD~\cite{madry} with step size
$\alpha{=}1/255$ and a default maximum step number $\eta{=}8$.
The parameter $u$ is equal to $\gamma$, much less than the loss upper bound
in order to avoid excessive hardness boost in $g_\mathsf{B}$ and $g_\mathsf{LGA}$.
Parameter $\lambda$ for $L_\text{ICS}$ is $0.5$ by default ($0.05$ on SOP).

The model performance on the benign (\ie, unperturbed) examples is measured in terms of 
Recall@1 (R@1), Recall@2 (R@2), mAP and NMI
following~\cite{revisiting,robrank}.
The adversarial robustness of a model is measured in Empirical Robustness Score
(ERS)~\cite{robrank}, a normalized score (the higher the better) from
a collection of (simplified white-box) attacks against DML, which are
optimized with PGD ($\eta=32$ for strong attack).
Since adversarial training is not ``gradient masking''~\cite{obfuscated}, the
performance of white-box attack can be regarded as the upper bound of the
black-box attacks, and thus a model that is empirically robust to the collection
of white-box attacks is expected to be robust in general.

Concretely, the collection of attacks for ERS include:
(1) CA+, CA-, QA+ and QA-~\cite{advrank}, which move some selected candidates
towards the topmost or bottommost part of ranking list;
(2) TMA~\cite{flowertower} which increases the cosine similarity between two arbitrary samples;
(3) ES~\cite{advrank,advdpqn}, which moves the embedding of a sample off its original position as
distant as possible;
(4) LTM~\cite{learn-to-misrank}, which perturbs the ranking result by minimizing the distance of
unmatched pairs while maximizing the distance of matched pairs;
(5) GTM~\cite{robrank}, which minimizes the distance between query and the
closest unmatching sample.
(6) GTT~\cite{robrank}, aims to move the top-$1$ candidate out of the top-$4$
retrieval results, which is simplified from \cite{qair}.
The setup of all the attacks for robustness evaluation is unchanged from
\cite{robrank} for fair comparison.
Further details of these attacks can be found in \cite{robrank}.

\subsection{Selection of Source \& Destination Hardness}
\label{sec:41}

\begin{table}
\resizebox{\columnwidth}{!}{
\setlength{\tabcolsep}{0.36em}
\begin{tabular}{c|cc|cc|cc|cc|cc}
	\toprule

	\multirow{2}{*}{\backslashbox{$H_\mathsf{S}$}{$H_\mathsf{D}$}} & \multicolumn{2}{c|}{Random} & \multicolumn{2}{c|}{Semihard} & \multicolumn{2}{c|}{Softhard} & \multicolumn{2}{c|}{Distance} & \multicolumn{2}{c}{Hardest}\tabularnewline
\cline{2-11} \cline{3-11} \cline{4-11} \cline{5-11} \cline{6-11} \cline{7-11} \cline{8-11} \cline{9-11} \cline{10-11} \cline{11-11} 
& R@1 & ERS & R@1 & ERS & R@1 & ERS & R@1 & ERS & R@1 & ERS\tabularnewline

\midrule

Random & \cellcolor{black!7}53.9 & \cellcolor{black!7}3.8 & 27.0 & 35.1 & \multicolumn{2}{c|}{Collapse} & \multicolumn{2}{c|}{Collapse} & \multicolumn{2}{c}{Collapse}\tabularnewline
Semihard & 43.9 & 5.4 & \cellcolor{black!7}44.0 & \cellcolor{black!7}5.0 & \multicolumn{2}{c|}{Collapse} & \multicolumn{2}{c|}{Collapse} & \multicolumn{2}{c}{Collapse}\tabularnewline
Softhard & 48.3 & 13.7 & 38.4 & 29.6 & \cellcolor{black!7}55.7 & \cellcolor{black!7}6.2 & \multicolumn{2}{c|}{Collapse} & \multicolumn{2}{c}{Collapse}\tabularnewline
Distance & 52.7 & 4.8 & 50.7 & 4.8 & \multicolumn{2}{c|}{Collapse} & \cellcolor{black!7}51.4 & \cellcolor{black!7}4.9 & 54.7 & 5.4\tabularnewline
Hardest & 51.0 & 4.7 & 52.2 & 4.8 & \multicolumn{2}{c|}{Collapse} & 52.6 & 5.1 & \cellcolor{black!7}48.9 & \cellcolor{black!7}5.0\tabularnewline

	\bottomrule
\end{tabular}
}
	\vspace{-0.5em}
\caption{Combinations of source \& destination hardness.
	Evaluated on the CUB Dataset with RN18 model.
	The last-epoch performance is reported instead of the peak performance for alignment.
	Models on the diagonal are regularly (instead of adversarially) trained.
}
	\label{tab:desth}

\end{table}

\begin{table*}
\resizebox{\linewidth}{!}{
\setlength{\tabcolsep}{0.48em}%
\renewcommand{\arraystretch}{1.05}

\begin{tabular}{c|cc|cccc|ccccc|ccccc|c}

	\toprule

\rowcolor{black!12} & & & \multicolumn{4}{c|}{\textbf{Benign Example}} & \multicolumn{10}{c|}{\textbf{White-Box Attacks for Robustness Evaluation}} & \tabularnewline
\cline{4-17} \cline{5-17} \cline{6-17} \cline{7-17} \cline{8-17} \cline{9-17} \cline{10-17} \cline{11-17} \cline{12-17} \cline{13-17} \cline{14-17} \cline{15-17} \cline{16-17} \cline{17-17}
\rowcolor{black!12}\multirow{-2}{*}{\textbf{Dataset}} & \multirow{-2}{*}{\textbf{Defense}} & \multirow{-2}{*}{$\eta$} & R@1$\uparrow$ & R@2$\uparrow$ & mAP$\uparrow$ & NMI$\uparrow$ & CA+$\uparrow$ & CA-$\downarrow$ & QA+$\uparrow$ & QA-$\downarrow$ & TMA$\downarrow$ & ES:D$\downarrow$ & ES:R$\uparrow$ & LTM$\uparrow$ & GTM$\uparrow$ & GTT$\uparrow$ & \multirow{-2}{*}{\textbf{ERS$\uparrow$}}\tabularnewline

	\midrule

CUB & N/A{[}$\mathcal{R}${]} & N/A & 53.9 & 66.4 & 26.1 & 59.5 & 0.0 & 100.0 & 0.0 & 99.9 & 0.883 & 1.762 & 0.0 & 0.0 & 14.1 & 0.0 & 3.8\tabularnewline
\hline
\multirow{5}{*}{CUB} & \textcolor{tab:cyan}{ACT{[}$\mathcal{R}${]}}\cite{robrank} & 2 & 46.5 & 58.4 & 29.1 & 55.6 & 0.6 & 98.9 & 0.4 & 98.1 & 0.837 & 1.666 & 0.2 & 0.2 & 19.6 & 0.0 & 5.8\tabularnewline
\rowcolor{black!7}\cellcolor{white} & \textcolor{tab:cyan}{ACT{[}$\mathcal{R}${]}}\cite{robrank} & 4 & 38.4 & 49.8 & 22.8 & 49.7 & 4.6 & 81.9 & 2.8 & 80.5 & 0.695 & 1.366 & 2.9 & 2.3 & 18.8 & 0.1 & 13.9\tabularnewline
 & \textcolor{tab:cyan}{ACT{[}$\mathcal{R}${]}}\cite{robrank} & 8 & 30.6 & 40.1 & 16.5 & 45.6 & 13.7 & 46.8 & 12.6 & 39.3 & 0.547 & 0.902 & 13.6 & 9.8 & 21.9 & 1.3 & 31.3\tabularnewline
\rowcolor{black!7}\cellcolor{white} & \textcolor{tab:cyan}{ACT{[}$\mathcal{R}${]}}\cite{robrank} & 16 & 28.6 & 38.7 & 15.1 & 43.7 & 15.8 & 37.9 & 16.0 & 31.5 & 0.496 & 0.834 & 11.3 & 9.8 & 21.2 & 2.1 & 34.7\tabularnewline
 & \textcolor{tab:cyan}{ACT{[}$\mathcal{R}${]}}\cite{robrank} & 32 & 27.5 & 38.2 & 12.2 & 43.0 & 15.5 & 37.7 & 15.1 & 32.2 & 0.472 & 0.821 & 11.1 & 9.4 & 14.9 & 1.0 & 33.9\tabularnewline
\hline
\multirow{5}{*}{CUB} & \textcolor{tab:orange}{ACT{[}$\mathcal{S}${]}}\cite{robrank} & 2 & 53.0 & 65.1 & 34.7 & 59.9 & 0.0 & 100.0 & 0.0 & 99.8 & 0.877 & 1.637 & 0.0 & 0.0 & 20.4 & 0.0 & 5.1\tabularnewline
\rowcolor{black!7}\cellcolor{white}& \textcolor{tab:orange}{ACT{[}$\mathcal{S}${]}}\cite{robrank} & 4 & 49.3 & 61.0 & 31.5 & 56.6 & 0.6 & 97.6 & 0.2 & 98.1 & 0.799 & 1.485 & 0.3 & 0.2 & 18.9 & 0.0 & 7.1\tabularnewline
& \textcolor{tab:orange}{ACT{[}$\mathcal{S}${]}}\cite{robrank} & 8 & 42.8 & 54.7 & 26.6 & 53.3 & 4.8 & 72.8 & 2.7 & 73.3 & 0.619 & 1.148 & 8.3 & 4.9 & 23.5 & 0.3 & 18.7\tabularnewline
\rowcolor{black!7}\cellcolor{white}& \textcolor{tab:orange}{ACT{[}$\mathcal{S}${]}}\cite{robrank} & 16 & 40.5 & 51.6 & 24.8 & 51.7 & 6.7 & 62.1 & 4.9 & 60.6 & 0.566 & 1.014 & 12.4 & 8.6 & 22.5 & 0.9 & 23.7\tabularnewline
& \textcolor{tab:orange}{ACT{[}$\mathcal{S}${]}}\cite{robrank} & 32 & 39.4 & 50.2 & 18.6 & 51.3 & 6.8 & 61.5 & 5.2 & 60.4 & 0.506 & 1.032 & 12.8 & 11.3 & 17.7 & 0.3 & 24.2\tabularnewline
\hline
\multirow{5}{*}{CUB} & \textcolor{tab:blue}{HM{[}$\mathcal{R},\mathcal{M}${]}} & 2 & 34.3 & 44.9 & 19.5 & 47.4 & 7.7 & 77.5 & 6.5 & 70.8 & 0.636 & 1.281 & 4.3 & 2.6 & 21.1 & 0.2 & 18.1\tabularnewline
\rowcolor{black!7}\cellcolor{white} & \textcolor{tab:blue}{HM{[}$\mathcal{R},\mathcal{M}${]}} & 4 & 30.7 & 40.3 & 16.4 & 45.3 & 13.9 & 60.4 & 13.5 & 48.1 & 0.582 & 1.041 & 6.6 & 6.6 & 20.2 & 1.2 & 27.1\tabularnewline
 & \textcolor{tab:blue}{HM{[}$\mathcal{R},\mathcal{M}${]}} & 8 & 27.0 & 36.0 & 13.2 & 42.5 & 19.4 & 48.0 & 22.2 & 32.0 & 0.535 & 0.867 & 11.6 & 10.4 & 19.3 & 2.9 & 35.1\tabularnewline
\rowcolor{black!7}\cellcolor{white} & \textcolor{tab:blue}{HM{[}$\mathcal{R},\mathcal{M}${]}} & 16 & 23.8 & 32.6 & 11.6 & 40.6 & 20.9 & 45.0 & 24.6 & 28.6 & 0.494 & 0.805 & 15.6 & 11.3 & 22.1 & 3.2 & 38.0\tabularnewline
 & \textcolor{tab:blue}{HM{[}$\mathcal{R},\mathcal{M}${]}} & 32 & 23.1 & 31.9 & 11.3 & 40.3 & 22.8 & 46.0 & 24.3 & 28.3 & 0.495 & 0.800 & 14.2 & 11.7 & 19.7 & 3.8 & 38.0\tabularnewline
\hline
\multirow{5}{*}{CUB} & \textcolor{tab:red}{HM{[}$\mathcal{S},\mathcal{M}${]}} & 2 & 44.5 & 56.1 & 27.8 & 53.3 & 1.9 & 87.7 & 1.6 & 88.8 & 0.827 & 1.101 & 3.7 & 0.3 & 19.0 & 0.0 & 11.6\tabularnewline
\rowcolor{black!7}\cellcolor{white} & \textcolor{tab:red}{HM{[}$\mathcal{S},\mathcal{M}${]}} & 4 & 40.6 & 51.8 & 24.2 & 51.0 & 7.3 & 64.1 & 6.3 & 60.9 & 0.715 & 0.894 & 7.9 & 4.4 & 22.8 & 0.2 & 22.1\tabularnewline
 & \textcolor{tab:red}{HM{[}$\mathcal{S},\mathcal{M}${]}} & 8 & 38.4 & 49.7 & 22.9 & 50.3 & 10.9 & 50.5 & 10.8 & 44.6 & 0.680 & 0.722 & 13.3 & 11.2 & 25.8 & 1.2 & 29.6\tabularnewline
\rowcolor{black!7}\cellcolor{white} & \textcolor{tab:red}{HM{[}$\mathcal{S},\mathcal{M}${]}} & 16 & 37.4 & 47.3 & 21.0 & 48.2 & 14.4 & 42.0 & 14.8 & 34.7 & 0.599 & 0.693 & 17.5 & 14.4 & 26.5 & 2.4 & 34.8\tabularnewline
 & \textcolor{tab:red}{HM{[}$\mathcal{S},\mathcal{M}${]}} & 32 & 35.3 & 46.1 & 20.2 & 48.0 & 15.1 & 41.8 & 15.2 & 33.0 & 0.589 & 0.686 & 18.7 & 14.9 & 27.8 & 2.9 & 35.7\tabularnewline

	\bottomrule

\end{tabular}

} \vspace{-0.5em}
\caption{Hardness manipulation in adversarial training.
The ``$\uparrow$'' mark means ``the higher the better'', while ``$\downarrow$''
	means the opposite.
	}
\label{tab:hmeff}
\end{table*}


As discussed in \cref{sec:31}, we start from the $H_\mathsf{D}$ calculated from
a harder benign triplet sampled by a different strategy, such as Random,
Semihard~\cite{facenet}, Softhard~\cite{revisiting},
Distance-weighted~\cite{distance} (\emph{abbr}., Distance), or the within-batch
Hardest negative sampling strategy, because we know these strategies do not
result in model collapse in regular training.


HM is flexible so that any existing or future triplet sampling strategy
can be used for the source triplet or calculating $H_\mathsf{D}$.
But not all potential combinations are expected to be effective for HM, as it
will not create an adversarial triplet when $H_\mathsf{S}\geqslant
H_\mathsf{D}$.
Thus, we sort the strategies based on the mean hardness of their
outputs in~\cref{tab:hsort}.
Then we adversarially train models on the CUB dataset with all combinations
respectively, and summarize their R@1 and ERS in \cref{tab:desth}.


For cases in the upper triangular of \cref{tab:desth} where $E[H_\mathsf{S}]
\leqslant E[H_\mathsf{D}]$, most of the given triplets will be turned
adversarial.
Although almost all of these cases end up with model collapse, the
$\text{HM}[\mathcal{R},\mathcal{M}]$ is still effective in improving the
robustness, with an expected performance drop in R@1.
The combination of Distance and Hardest triplets does not trigger model
collapse due to the small $E[H_\mathsf{D}-H_\mathsf{S}]$, which leads to weak
adversarial examples and a negligible robustness gain.

For cases in lower triangular of \cref{tab:desth}, where $E[H_\mathsf{S}]
\geqslant E[H_\mathsf{D}]$, a large portion of given triplets will be unchanged
according to \cref{eq:hm}, and hence lead to weak robustness.
As an exception, $\text{HM}[\mathcal{S},\mathcal{M}]$ is still effective in
improving adversarial robustness, where a notable number of adversarial
examples are created due the high $\text{Var}[H]$ of Softhard.
Although $E[H]$ of Softhard is less than that of Distance or Hardest, some hard
adversarial examples are still created%
\footnote{Differently, Softhard also samples a
hard positive instead of a random positive besides a hard negative.
As a result, the hardness of a small number of Softhard triplets will be 
greater than that of a given Hardest triplet.}
due to its large $\text{Var}[H]$, which
still result in a slow collapse.


In practice, HM creates mini-batches mixing some unperturbed source triplets and
some adversarial triplets.
The $\text{HM}[\mathcal{R}, \mathcal{M}]$ and $\text{HM}[\mathcal{S},
\mathcal{M}]$ achieve such balanced mixtures.
Subsequent experiments will be based on the two effective combinations.
Empirically, the hardness range of Semihard strategy,
\ie, $[-\gamma,0]$ is found appropriate for $H_\mathsf{D}$.

\subsection{Effectiveness of Our Approach}
\label{sec:42}

\begin{figure}[t]
	\includegraphics[width=\columnwidth]{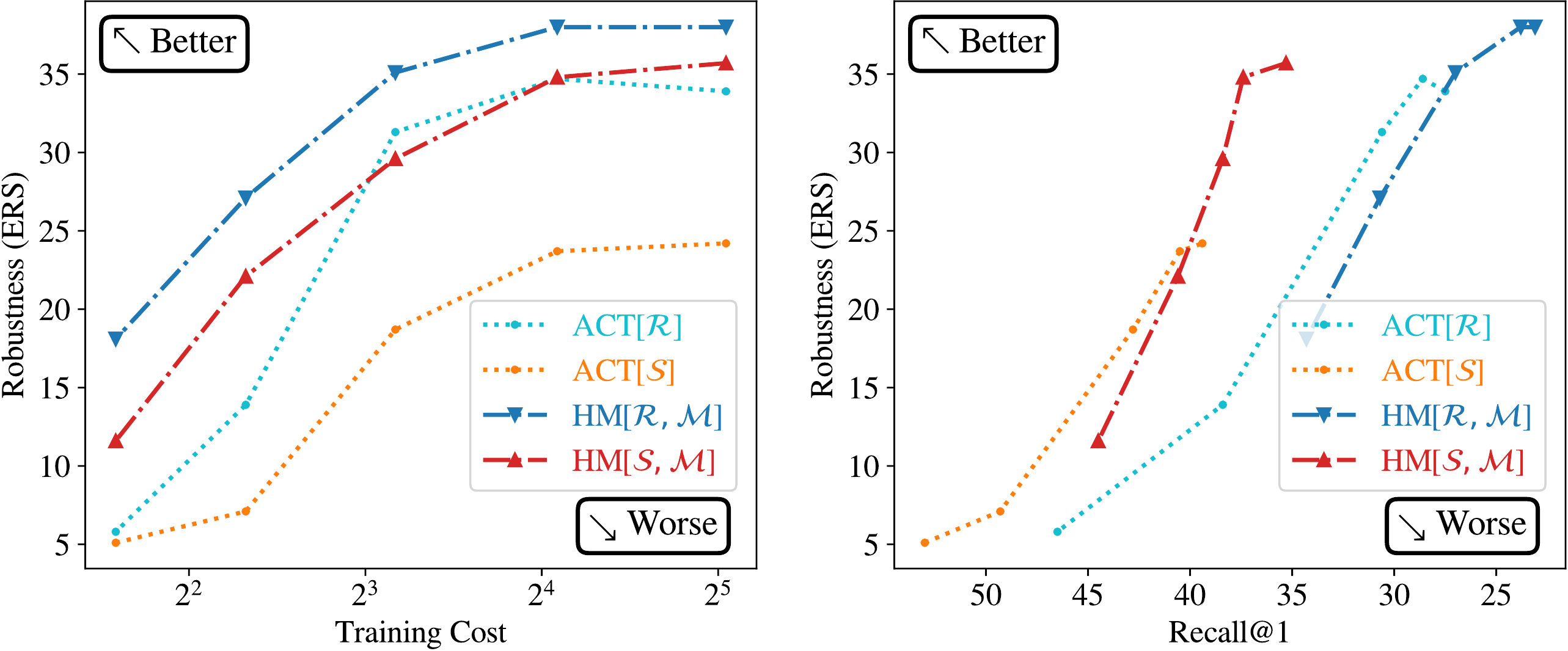}
	\vspace{-1.5em}
	\caption{Performance of $\text{HM}[\mathcal{R},\mathcal{M}]$
	\& $\text{HM}[\mathcal{S},\mathcal{M}]$ in \cref{tab:hmeff}.
	}
	\label{fig:hmeff}
\end{figure}

\noindent\textbf{I. Hardness Manipulation.}
To validate HM with $H_\mathsf{D}$ calculated from benign
triplets, we adversarially train models using $\text{HM}[\mathcal{R},
\mathcal{M}]$ and $\text{HM}[\mathcal{S}, \mathcal{M}]$ on the CUB dataset,
with varying PGD steps, \ie, $\eta\in\{2,4,8,16,32\}$, respectively.
The results can be found in \cref{tab:hmeff}.
The performance of the state-of-the-art defense, \ie, ACT~\cite{robrank} is
provided as a baseline.
ACT[$\mathcal{R}$] and ACT[$\mathcal{S}$] mean the training triplet is sampled
using Random and Softhard strategy, respectively.
We also plot curves in \cref{fig:hmeff} based on the robustness, training
cost\footnote{Training cost is the number of times for forward and backward
propagation in each adversarial training iteration, which is calculated as
$\eta+1$.}, as well as the R@1 performance on benign examples.


As shown, ACT[$\mathcal{R}$] can achieve a high ERS, but with a significant
performance drop in R@1, while ACT[$\mathcal{S}$] can retain a relative high
R@1, but is much less efficient in gaining robustness under a fixed training
cost.
Notably, ACT relies on the attack that can successfully pull the adversarial
positive and negative samples close to each other in order to learn robust
features~\cite{robrank}.
As a result, ACT's ERS with a small $\eta$ (indicating a weak attack effect) is
relatively low.

In contrast, HM[$\mathcal{R},\mathcal{M}$] achieves an even higher ERS under
the same training cost, but with a larger penalty in R@1 compared to
ACT[$\mathcal{R}$].
Compared to ACT[$\mathcal{S}$], HM[$\mathcal{S},\mathcal{M}$] is able to retain
a relatively high R@1, but in a much higher efficiency.
As can be seen from \cref{fig:hmeff}, HM[$\mathcal{R},\mathcal{M}$] achieves the
highest ERS and efficiency but with the most significant drop in R@1, which is
not acceptable in applications.
Apart from that, HM[$\mathcal{S},\mathcal{M}$] achieves a promising result
in every aspect.
Its efficiency in gaining robustness is basically on par with
ACT[$\mathcal{R}$], but can achieve a significantly higher R@1.
It achieves a balance between ERS and R@1 on par with ACT[$\mathcal{S}$], but
in a significantly higher efficiency.

\begin{table*}
\resizebox{\linewidth}{!}{
\setlength{\tabcolsep}{0.48em}
\renewcommand{\arraystretch}{1.02}%

\begin{tabular}{c|cc|cccc|ccccc|ccccc|c}

	\toprule

\rowcolor{black!12} & & & \multicolumn{4}{c|}{\textbf{Benign Example}} & \multicolumn{10}{c|}{\textbf{White-Box Attacks for Robustness Evaluation}} & \tabularnewline
\cline{4-17} \cline{5-17} \cline{6-17} \cline{7-17} \cline{8-17} \cline{9-17} \cline{10-17} \cline{11-17} \cline{12-17} \cline{13-17} \cline{14-17} \cline{15-17} \cline{16-17} \cline{17-17}
\rowcolor{black!12}\multirow{-2}{*}{\textbf{Dataset}} &  \multirow{-2}{*}{\textbf{Defense}} & \multirow{-2}{*}{$\eta$}  & R@1$\uparrow$ & R@2$\uparrow$ & mAP$\uparrow$ & NMI$\uparrow$ & CA+$\uparrow$ & CA-$\downarrow$ & QA+$\uparrow$ & QA-$\downarrow$ & TMA$\downarrow$ & ES:D$\downarrow$ & ES:R$\uparrow$ & LTM$\uparrow$ & GTM$\uparrow$ & GTT$\uparrow$ & \multirow{-2}{*}{\textbf{ERS$\uparrow$}}\tabularnewline

	\midrule

& \textcolor{tab:red}{HM{[}$\mathcal{S},\mathcal{M}${]}} & 8 & 38.4 & 49.7 & 22.9 & 50.3 & 10.9 & 50.5 & 10.8 & 44.6 & 0.680 & 0.722 & 13.3 & 11.2 & 25.8 & 1.2 & 29.6\tabularnewline
	\cline{2-18}
 & HM{[}$\mathcal{S},g_{\mathsf{B}}(\mathcal{M})${]} ($\xi=0.1$) & 8 & 36.5 & 48.0 & 21.4 & 48.4 & 13.0 & 44.0 & 13.2 & 35.6 & 0.667 & 0.628 & 20.3 & 13.2 & 26.7 & 2.8 & 33.8\tabularnewline
\rowcolor{black!7}\cellcolor{white} & HM{[}$\mathcal{S},0${]} & 8 & 0.8 & 0.9 & 0.8 & 6.0 & 19.8 & 92.4 & 42.0 & 51.9 & 1.000 & 0.000 & 1.2 & 1.2 & 1.0 & 14.1 & 29.7\tabularnewline
 & HM{[}$\mathcal{S},-\gamma/2${]} & 8 & 36.8 & 47.9 & 21.7 & 48.5 & 12.7 & 41.5 & 12.2 & 35.7 & 0.668 & 0.633 & 18.1 & 14.3 & 28.4 & 2.9 & 33.8\tabularnewline
\rowcolor{black!7}\cellcolor{white} & HM{[}$\mathcal{S},-\gamma${]} & 8 & 37.8 & 48.1 & 22.1 & 48.7 & 11.7 & 48.4 & 11.3 & 43.2 & 0.541 & 0.850 & 15.2 & 11.6 & 26.1 & 1.3 & 31.2\tabularnewline
\multirow{-6}{*}{CUB}  & \textcolor{tab:pink}{HM{[}$\mathcal{S},g_{\mathsf{LGA}}${]}} & 8 & 38.0 & 48.3 & 21.8 & 49.3 & 12.7 & 46.4 & 11.6 & 39.9 & 0.567 & 0.783 & 16.8 & 11.9 & 27.9 & 1.4 & 32.4\tabularnewline

	\hline

\multirow{5}{*}{CUB} & HM{[}$\mathcal{S},-\gamma/2${]} & 2 & 44.5 & 55.9 & 27.6 & 53.3 & 2.4 & 86.1 & 1.3 & 87.7 & 0.809 & 1.091 & 1.5 & 1.8 & 22.1 & 0.1 & 12.4\tabularnewline
\rowcolor{black!7}\cellcolor{white} & HM{[}$\mathcal{S},-\gamma/2${]} & 4 & 40.0 & 50.7 & 23.8 & 50.3 & 8.0 & 59.8 & 7.5 & 55.5 & 0.694 & 0.860 & 10.2 & 6.5 & 26.2 & 0.4 & 24.7\tabularnewline
 & HM{[}$\mathcal{S},-\gamma/2${]} & 8 & 36.8 & 47.9 & 21.7 & 48.5 & 12.7 & 41.5 & 12.2 & 35.7 & 0.668 & 0.633 & 18.1 & 14.3 & 28.4 & 2.9 & 33.8\tabularnewline
\rowcolor{black!7}\cellcolor{white} & HM{[}$\mathcal{S},-\gamma/2${]} & 16 & 35.2 & 45.8 & 20.3 & 47.6 & 15.0 & 36.3 & 15.1 & 30.7 & 0.638 & 0.595 & 19.6 & 16.8 & 28.7 & 3.5 & 36.8\tabularnewline
 & HM{[}$\mathcal{S},-\gamma/2${]} & 32 & 34.7 & 45.5 & 20.0 & 47.5 & 15.0 & 36.7 & 15.1 & 29.9 & 0.631 & 0.611 & 20.1 & 17.2 & 29.3 & 3.5 & 37.0\tabularnewline
\hline
\multirow{5}{*}{CUB} & \textcolor{tab:pink}{HM{[}$\mathcal{S},g_{\mathsf{LGA}}${]}} & 2 & 47.5 & 59.3 & 30.1 & 55.3 & 1.8 & 88.1 & 1.1 & 88.9 & 0.854 & 1.022 & 2.3 & 0.8 & 21.2 & 0.0 & 11.7\tabularnewline
\rowcolor{black!7}\cellcolor{white} & \textcolor{tab:pink}{HM{[}$\mathcal{S},g_{\mathsf{LGA}}${]}} & 4 & 42.7 & 53.6 & 26.3 & 52.6 & 6.5 & 67.3 & 4.6 & 65.0 & 0.734 & 0.893 & 6.6 & 5.8 & 23.7 & 0.3 & 20.8\tabularnewline
 & \textcolor{tab:pink}{HM{[}$\mathcal{S},g_{\mathsf{LGA}}${]}} & 8 & 38.0 & 48.3 & 21.8 & 49.3 & 12.7 & 46.4 & 11.6 & 39.9 & 0.567 & 0.783 & 16.8 & 11.9 & 27.9 & 1.4 & 32.4\tabularnewline
\rowcolor{black!7}\cellcolor{white} & \textcolor{tab:pink}{HM{[}$\mathcal{S},g_{\mathsf{LGA}}${]}} & 16 & 37.0 & 47.2 & 21.3 & 48.4 & 13.6 & 42.2 & 13.1 & 35.9 & 0.533 & 0.757 & 16.3 & 15.3 & 27.2 & 2.1 & 34.5\tabularnewline
 & \textcolor{tab:pink}{HM{[}$\mathcal{S},g_{\mathsf{LGA}}${]}} & 32 & 36.5 & 46.7 & 21.0 & 48.6 & 14.7 & 39.6 & 15.6 & 34.2 & 0.523 & 0.736 & 16.5 & 15.0 & 26.7 & 2.9 & 35.9\tabularnewline
 
	\bottomrule

\end{tabular}

	}
	\vspace{-0.8em}
\caption{Effectiveness of gradual adversary as $H_\mathsf{D}$ in hardness manipulation.}
\label{tab:gaeff}
\end{table*}

\begin{table*}
	\vspace{-1em}
\resizebox{\linewidth}{!}{
\setlength{\tabcolsep}{0.48em}
\renewcommand{\arraystretch}{1.02}%

\begin{tabular}{c|cc|cccc|ccccc|ccccc|c}

	\toprule

\rowcolor{black!12} & &  & \multicolumn{4}{c|}{\textbf{Benign Example}} & \multicolumn{10}{c|}{\textbf{White-Box Attacks for Robustness Evaluation}} & \tabularnewline
\cline{4-17} \cline{5-17} \cline{6-17} \cline{7-17} \cline{8-17} \cline{9-17} \cline{10-17} \cline{11-17} \cline{12-17} \cline{13-17} \cline{14-17} \cline{15-17} \cline{16-17} \cline{17-17}
\rowcolor{black!12}\multirow{-2}{*}{\textbf{Dataset}} & \multirow{-2}{*}{\textbf{Defense}} & \multirow{-2}{*}{$\eta$} & R@1$\uparrow$ & R@2$\uparrow$ & mAP$\uparrow$ & NMI$\uparrow$ & CA+$\uparrow$ & CA-$\downarrow$ & QA+$\uparrow$ & QA-$\downarrow$ & TMA$\downarrow$ & ES:D$\downarrow$ & ES:R$\uparrow$ & LTM$\uparrow$ & GTM$\uparrow$ & GTT$\uparrow$ & \multirow{-2}{*}{\textbf{ERS$\uparrow$}}\tabularnewline

	\midrule

& \textcolor{tab:blue}{HM{[}$\mathcal{R},\mathcal{M}${]}} & 8 & 27.0 & 36.0 & 13.2 & 42.5 & 19.4 & 48.0 & 22.2 & 32.0 & 0.535 & 0.867 & 11.6 & 10.4 & 19.3 & 2.9 & 35.1\tabularnewline
\rowcolor{black!7}\cellcolor{white} \multirow{-2}{*}{CUB} & HM{[}$\mathcal{R},\mathcal{M}${]}\&ICS & 8 & 25.6 & 34.3 & 12.5 & 41.8 & 21.9 & 41.0 & 23.6 & 26.4 & 0.497 & 0.766 & 14.5 & 13.0 & 21.8 & 4.7 & 39.0\tabularnewline
\hline 
& \textcolor{tab:red}{HM{[}$\mathcal{S},\mathcal{M}${]}} & 8 & 38.4 & 49.7 & 22.9 & 50.3 & 10.9 & 50.5 & 10.8 & 44.6 & 0.680 & 0.722 & 13.3 & 11.2 & 25.8 & 1.2 & 29.6\tabularnewline
\rowcolor{black!7}\cellcolor{white}\multirow{-2}{*}{CUB}  & HM{[}$\mathcal{S},\mathcal{M}${]}\&ICS & 8 & 36.9 & 48.9 & 21.6 & 48.8 & 12.4 & 42.9 & 12.5 & 36.6 & 0.850 & 0.446 & 17.0 & 13.9 & 27.2 & 1.9 & 32.3\tabularnewline
\hline 
& HM{[}$\mathcal{R},g_{\mathsf{LGA}}${]} & 8 & 24.8 & 33.9 & 12.2 & 41.6 & 21.4 & 45.0 & 21.7 & 31.3 & 0.452 & 0.846 & 13.2 & 12.0 & 20.9 & 4.6 & 37.3\tabularnewline
\rowcolor{black!7}\cellcolor{white}\multirow{-2}{*}{CUB}  & HM{[}$\mathcal{R},g_{\mathsf{LGA}}${]}\&ICS & 8 & 25.7 & 35.2 & 12.8 & 41.7 & 22.1 & 37.1 & 23.4 & 23.7 & 0.464 & 0.725 & 14.5 & 13.3 & 21.1 & 5.3 & 40.2\tabularnewline
\hline 
& \textcolor{tab:pink}{HM{[}$\mathcal{S},g_{\mathsf{LGA}}${]}} & 8 & 38.0 & 48.3 & 21.8 & 49.3 & 12.7 & 46.4 & 11.6 & 39.9 & 0.567 & 0.783 & 16.8 & 11.9 & 27.9 & 1.4 & 32.4\tabularnewline
\rowcolor{black!7}\cellcolor{white} & \textcolor{crimson}{HM{[}$\mathcal{S},g_{\mathsf{LGA}}${]}\&ICS} & 8 & 37.2 & 47.8 & 21.4 & 48.4 & 12.9 & 40.9 & 14.7 & 33.7 & 0.806 & 0.487 & 17.1 & 13.2 & 26.3 & 2.3 & 33.5\tabularnewline
\multirow{-3}{*}{CUB} & HM{[}$\mathcal{S},g_{\mathsf{LGA}}${]}\&ICS($\lambda{=}1.0$) & 8 & 36.0 & 46.7 & 20.7 & 48.0 & 14.2 & 41.0 & 15.1 & 31.7 & 0.907 & 0.329 & 17.0 & 14.2 & 24.5 & 2.1 & 33.7\tabularnewline
\hline 
\rowcolor{black!7}\cellcolor{white}& \textcolor{crimson}{HM{[}$\mathcal{S},g_{\mathsf{LGA}}${]}\&ICS} & 2 & 45.2 & 57.2 & 28.5 & 53.7 & 3.0 & 79.9 & 2.4 & 78.9 & 0.936 & 0.609 & 3.6 & 1.2 & 19.9 & 0.0 & 15.2\tabularnewline
 & \textcolor{crimson}{HM{[}$\mathcal{S},g_{\mathsf{LGA}}${]}\&ICS} & 4 & 41.8 & 53.0 & 25.3 & 52.0 & 8.1 & 57.3 & 7.9 & 54.1 & 0.892 & 0.514 & 9.8 & 6.7 & 22.9 & 0.5 & 24.6\tabularnewline
\rowcolor{black!7}\cellcolor{white} & \textcolor{crimson}{HM{[}$\mathcal{S},g_{\mathsf{LGA}}${]}\&ICS} & 8 & 37.2 & 47.8 & 21.4 & 48.4 & 12.9 & 40.9 & 14.7 & 33.7 & 0.806 & 0.487 & 17.1 & 13.2 & 26.3 & 2.3 & 33.5\tabularnewline
 & \textcolor{crimson}{HM{[}$\mathcal{S},g_{\mathsf{LGA}}${]}\&ICS} & 16 & 35.5 & 46.4 & 20.4 & 47.5 & 14.9 & 37.2 & 17.1 & 30.3 & 0.771 & 0.495 & 18.2 & 15.3 & 28.7 & 2.8 & 36.0\tabularnewline
\rowcolor{black!7}\cellcolor{white}\multirow{-5}{*}{CUB}  & \textcolor{crimson}{HM{[}$\mathcal{S},g_{\mathsf{LGA}}${]}\&ICS} & 32 & 34.9 & 45.0 & 19.8 & 47.1 & 15.5 & 37.7 & 16.6 & 30.9 & 0.753 & 0.506 & 17.9 & 16.7 & 27.3 & 2.9 & 36.0\tabularnewline

	\bottomrule

\end{tabular}

	}
	\vspace{-0.8em}
\caption{Intra-class structure loss in conjunction with hardness manipulation for adversarial training of a DML Model.}
\label{tab:ics}
\end{table*}


Overall, as discussed in \cref{sec:31}, HM uses the same projected gradient as
to directly maximize the hardness, which endows this method a high efficiency
in creating strong adversarial examples at a fixed training cost.
Besides, unlike ACT, HM does not rely on the attack to successfully move the
embeddings to some specific locations, and hence does not suffer from low
efficiency when $\eta$ is small.
HM[$\mathcal{R},\mathcal{M}$] creates training batches with some Random benign
examples and a large portion of Semihard adversarial examples, and hence
achieve a high ERS and a relatively low R@1 because the Random sampling
strategy is not selective to benign samples on which the model does not generalize well.
HM[$\mathcal{S},\mathcal{M}$] creates training batches with some Semihard
adversarial examples and a large portion of Softhard benign examples, and hence
achieve a relatively high ERS and a high R@1 because Softhard sampling
strategy is selective.
Further experiments will be based on HM[$\mathcal{S},\mathcal{M}$].

\begin{figure}[t]
	\vspace{-1.5em}
	\includegraphics[width=\columnwidth]{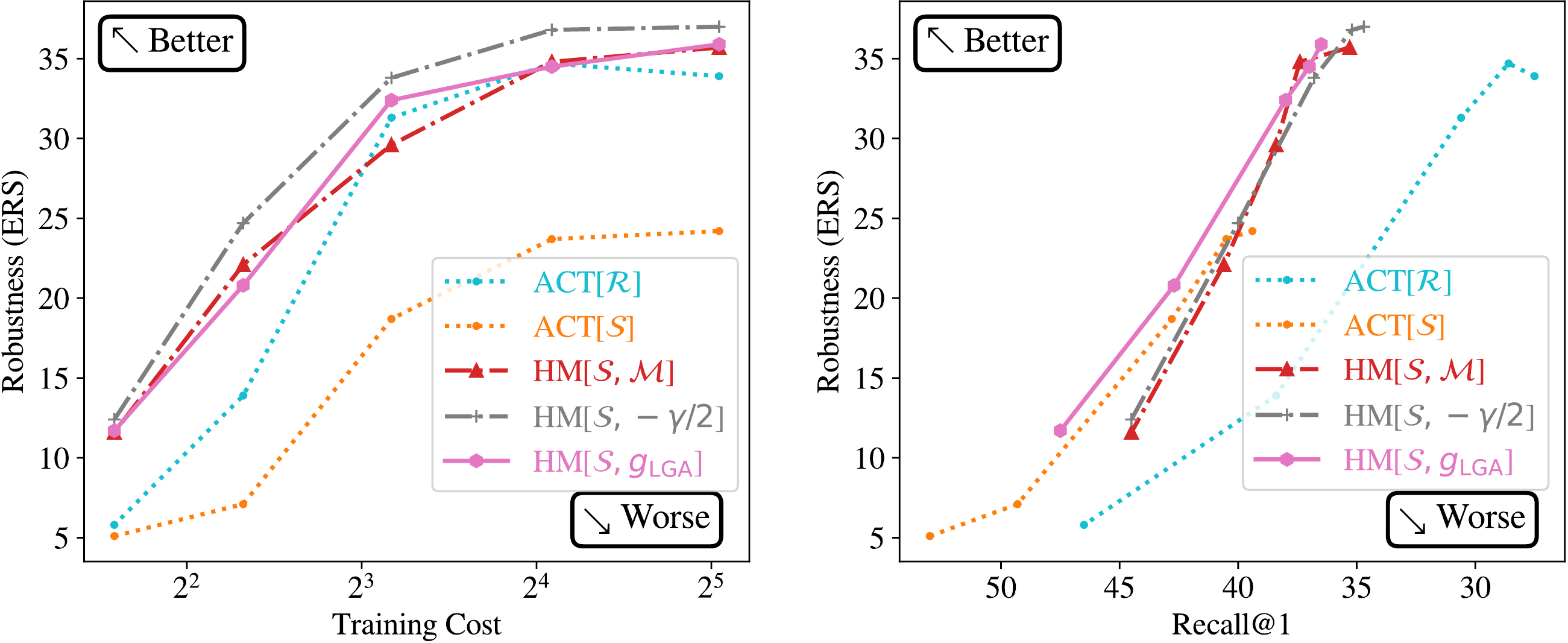}
	\vspace{-1.9em}
	\caption{Performance of ``HM[$\mathcal{S},g_\mathsf{LGA}$]'' in
	\cref{tab:gaeff}.}
	\vspace{-1em}
	\label{fig:gaeff}
\end{figure}

\noindent\textbf{II. Gradual Adversary.}
HM[$\mathcal{S},\mathcal{M}$] may still
suffer from the imbalance between learning the embeddings and gaining
adversarial robustness as discussed in \cref{sec:32}.
Hence, we conduct further experiments 
following the discussion, as shown in \cref{tab:gaeff} and \cref{fig:gaeff}.
Compared to HM[$\mathcal{S},\mathcal{M}$], slightly boosting the hardness
with $g_\mathsf{B}(\cdot)$ benefits the ERS, but
results in a notably lower R@1;
A constant $H_\mathsf{D}$ at the upper bound of Semihard (\ie, $0$; too high
for both the early and the late phase of training) renders model collapse;
$H_\mathsf{D}$ at the lower bound (\ie, $-\gamma$; too low for the late phase)
leads to insignificant ERS improvement;
$H_\mathsf{D}{=}-\gamma/2$ provides a fair balance in training objectives, but
still suffers from inflexibility.
In contrast, being not susceptible to the mentioned problems of other choices,
HM[$\mathcal{S},g_\mathsf{LGA}$] achieves an ERS on par with
HM[$\mathcal{S},\mathcal{M}$], but is at the lowest R@1 performance penalty among all choices.
Its ERS is marginally lower than HM[$\mathcal{S},{-}\gamma/2$] because the
observed loss value converges around ${-}\gamma/2$ due to optimization
difficulty, which means
adversarial triplets with $H_\mathsf{D}{\in}[{-}\gamma/2,0]$
are seldom created.

\begin{figure}[t]
	\vspace{-1.5em}
	\includegraphics[width=\columnwidth]{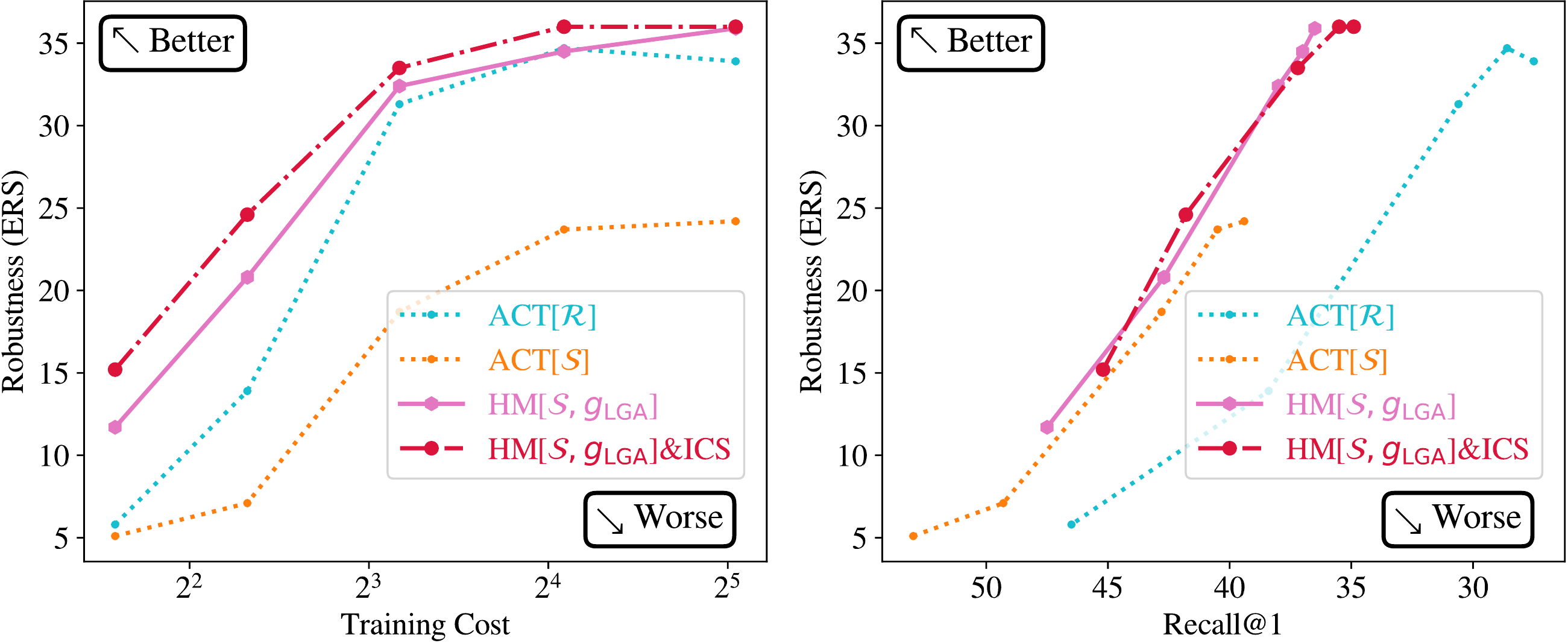}
	\vspace{-1.9em}
	\caption{Performance of ``HM[$\mathcal{S},g_\mathsf{LGA}$]\&ICS'' in \cref{tab:ics}.}
	\vspace{-1em}
	\label{fig:icscurve}
\end{figure}

\noindent\textbf{III. Intra-Class Structure.}
$L_\text{ICS}$ is independent to HM, but is incompatible with ACT as it does
not create adversarial anchor.
Thus, we validate this loss term with HM.
As shown in \cref{tab:ics} and \cref{fig:icscurve}, $L_\text{ICS}$ consistently
leads to a higher efficiency in gaining higher robustness at a low training
cost, while retaining an acceptable trade-off in R@1.

\noindent\textbf{IV. Summary.}
Eventually, HM[$\mathcal{S},g_\mathsf{LGA}$]\&ICS outperforms
the state-of-the-art defense in robustness, training efficiency, and R@1 
performance, as shown in \cref{fig:introplot}.

\begin{table*}
\resizebox{\linewidth}{!}{
\setlength{\tabcolsep}{0.47em}
\renewcommand{\arraystretch}{1.05}%

\begin{tabular}{c|cc|cccc|ccccc|ccccc|c}

	\toprule

\rowcolor{black!12} & & & \multicolumn{4}{c|}{\textbf{Benign Example}} & \multicolumn{10}{c|}{\textbf{White-Box Attacks for Robustness Evaluation}} & \tabularnewline
\cline{4-17} \cline{5-17} \cline{6-17} \cline{7-17} \cline{8-17} \cline{9-17} \cline{10-17} \cline{11-17} \cline{12-17} \cline{13-17} \cline{14-17} \cline{15-17} \cline{16-17} \cline{17-17}
\rowcolor{black!12}\multirow{-2}{*}{\textbf{Dataset}} & \multirow{-2}{*}{\textbf{Defense}} & \multirow{-2}{*}{$\eta$} & R@1$\uparrow$ & R@2$\uparrow$ & mAP$\uparrow$ & NMI$\uparrow$ & CA+$\uparrow$ & CA-$\downarrow$ & QA+$\uparrow$ & QA-$\downarrow$ & TMA$\downarrow$ & ES:D$\downarrow$ & ES:R$\uparrow$ & LTM$\uparrow$ & GTM$\uparrow$ & GTT$\uparrow$ & \multirow{-2}{*}{\textbf{ERS$\uparrow$}}\tabularnewline

	\midrule

& N/A{[}$\mathcal{R}${]} & N/A & 53.9 & 66.4 & 26.1 & 59.5 & 0.0 & 100.0 & 0.0 & 99.9 & 0.883 & 1.762 & 0.0 & 0.0 & 14.1 & 0.0 & 3.8\tabularnewline
\cline{2-18} \cline{3-18} \cline{4-18} \cline{5-18} \cline{6-18} \cline{7-18} \cline{8-18} \cline{9-18} \cline{10-18} \cline{11-18} \cline{12-18} \cline{13-18} \cline{14-18} \cline{15-18} \cline{16-18} \cline{17-18} \cline{18-18} 
 & EST{[}$\mathcal{R}${]}~\cite{advrank} & 8 & 37.1 & 47.3 & 20.0 & 46.4 & 0.5 & 97.3 & 0.5 & 91.3 & 0.875 & 1.325 & 3.9 & 0.4 & 14.9 & 0.0 & 7.9\tabularnewline
\rowcolor{black!7}\cellcolor{white} & \textcolor{cyan}{ACT{[}$\mathcal{R}${]}}~\cite{robrank} & 8 & 30.6 & 40.1 & 16.5 & 45.6 & 13.7 & 46.8 & 12.6 & 39.3 & 0.547 & 0.902 & 13.6 & 9.8 & 21.9 & 1.3 & 31.3\tabularnewline
& \textcolor{tab:pink}{HM{[}$\mathcal{S},g_{\mathsf{LGA}}${]}} & 8 & 38.0 & 48.3 & 21.8 & 49.3 & 12.7 & 46.4 & 11.6 & 39.9 & 0.567 & 0.783 & 16.8 & 11.9 & 27.9 & 1.4 & 32.4\tabularnewline
\rowcolor{black!7}\cellcolor{white} & \textcolor{crimson}{HM{[}$\mathcal{S},g_{\mathsf{LGA}}${]}\&ICS} & 8 & 37.2 & 47.8 & 21.4 & 48.4 & 12.9 & 40.9 & 14.7 & 33.7 & 0.806 & 0.487 & 17.1 & 13.2 & 26.3 & 2.3 & 33.5\tabularnewline
\cline{2-18} \cline{3-18} \cline{4-18} \cline{5-18} \cline{6-18} \cline{7-18} \cline{8-18} \cline{9-18} \cline{10-18} \cline{11-18} \cline{12-18} \cline{13-18} \cline{14-18} \cline{15-18} \cline{16-18} \cline{17-18} \cline{18-18} 
 & EST{[}$\mathcal{R}${]}~\cite{advrank} & 32 & 8.5 & 13.0 & 2.6 & 25.2 & 2.7 & 97.9 & 0.4 & 97.3 & 0.848 & 1.576 & 1.4 & 0.0 & 4.0 & 0.0 & 5.3\tabularnewline
\rowcolor{black!7}\cellcolor{white} & \textcolor{cyan}{ACT{[}$\mathcal{R}${]}}~\cite{robrank} & 32 & 27.5 & 38.2 & 12.2 & 43.0 & 15.5 & 37.7 & 15.1 & 32.2 & 0.472 & 0.821 & 11.1 & 9.4 & 14.9 & 1.0 & 33.9\tabularnewline
& \textcolor{tab:pink}{HM{[}$\mathcal{S},g_{\mathsf{LGA}}${]}} & 32 & 36.5 & 46.7 & 21.0 & 48.6 & 14.7 & 39.6 & 15.6 & 34.2 & 0.523 & 0.736 & 16.5 & 15.0 & 26.7 & 2.9 & 35.9\tabularnewline
\rowcolor{black!7}\cellcolor{white}\multirow{-9}{*}{CUB} & \textcolor{crimson}{HM{[}$\mathcal{S},g_{\mathsf{LGA}}${]}\&ICS} & 32 & 34.9 & 45.0 & 19.8 & 47.1 & 15.5 & 37.7 & 16.6 & 30.9 & 0.753 & 0.506 & 17.9 & 16.7 & 27.3 & 2.9 & 36.0\tabularnewline

	\midrule 

& N/A{[}$\mathcal{R}${]} & N/A & 62.5 & 74.0 & 23.8 & 57.0 & 0.2 & 100.0 & 0.1 & 99.6 & 0.874 & 1.816 & 0.0 & 0.0 & 13.4 & 0.0 & 3.6\tabularnewline
\cline{2-18} \cline{3-18} \cline{4-18} \cline{5-18} \cline{6-18} \cline{7-18} \cline{8-18} \cline{9-18} \cline{10-18} \cline{11-18} \cline{12-18} \cline{13-18} \cline{14-18} \cline{15-18} \cline{16-18} \cline{17-18} \cline{18-18} 
 & EST{[}$\mathcal{R}${]}~\cite{advrank} & 8 & 57.1 & 68.4 & 30.3 & 47.7 & 0.1 & 99.9 & 0.1 & 98.1 & 0.902 & 1.681 & 0.7 & 0.2 & 15.4 & 0.0 & 4.4\tabularnewline
\rowcolor{black!7}\cellcolor{white} & \textcolor{cyan}{ACT{[}$\mathcal{R}${]}}~\cite{robrank} & 8 & 46.8 & 58.0 & 23.4 & 45.5 & 19.3 & 33.1 & 20.3 & 32.3 & 0.413 & 0.760 & 18.4 & 15.0 & 28.6 & 1.2 & 39.8\tabularnewline
& \textcolor{tab:pink}{HM{[}$\mathcal{S},g_{\mathsf{LGA}}${]}} & 8 & 63.2 & 73.7 & 36.8 & 53.5 & 15.3 & 32.0 & 17.9 & 33.9 & 0.463 & 0.653 & 23.4 & 28.5 & 44.6 & 5.8 & 42.4\tabularnewline
\rowcolor{black!7}\cellcolor{white} & \textcolor{crimson}{HM{[}$\mathcal{S},g_{\mathsf{LGA}}${]}\&ICS} & 8 & 61.7 & 72.6 & 35.5 & 51.8 & 21.0 & 23.3 & 23.1 & 22.2 & 0.698 & 0.415 & 31.2 & 38.0 & 47.8 & 9.6 & 47.9\tabularnewline
\cline{2-18} \cline{3-18} \cline{4-18} \cline{5-18} \cline{6-18} \cline{7-18} \cline{8-18} \cline{9-18} \cline{10-18} \cline{11-18} \cline{12-18} \cline{13-18} \cline{14-18} \cline{15-18} \cline{16-18} \cline{17-18} \cline{18-18} 
 & EST{[}$\mathcal{R}${]}~\cite{advrank} & 32 & 30.7 & 41.0 & 5.6 & 31.8 & 1.2 & 98.1 & 0.4 & 91.8 & 0.880 & 1.281 & 2.9 & 0.7 & 8.2 & 0.0 & 7.3\tabularnewline
\rowcolor{black!7}\cellcolor{white} & \textcolor{cyan}{ACT{[}$\mathcal{R}${]}}~\cite{robrank} & 32 & 43.4 & 54.6 & 11.8 & 42.9 & 18.0 & 32.3 & 17.5 & 30.5 & 0.383 & 0.763 & 16.3 & 15.3 & 20.7 & 1.6 & 38.6\tabularnewline
& \textcolor{tab:pink}{HM{[}$\mathcal{S},g_{\mathsf{LGA}}${]}} & 32 & 62.3 & 72.5 & 35.3 & 52.7 & 17.4 & 28.2 & 18.2 & 28.8 & 0.426 & 0.613 & 27.1 & 30.7 & 42.3 & 7.9 & 44.9\tabularnewline
\rowcolor{black!7}\cellcolor{white}\multirow{-9}{*}{CARS} & \textcolor{crimson}{HM{[}$\mathcal{S},g_{\mathsf{LGA}}${]}\&ICS} & 32 & 60.2 & 71.6 & 33.9 & 51.2 & 19.3 & 25.9 & 19.6 & 25.7 & 0.650 & 0.446 & 30.3 & 36.7 & 46.0 & 8.8 & 46.0\tabularnewline

	\midrule

& N/A{[}$\mathcal{R}${]} & N/A & 62.9 & 68.5 & 39.2 & 87.4 & 0.1 & 99.3 & 0.2 & 99.1 & 0.845 & 1.685 & 0.0 & 0.0 & 6.3 & 0.0 & 4.0\tabularnewline
\cline{2-18} \cline{3-18} \cline{4-18} \cline{5-18} \cline{6-18} \cline{7-18} \cline{8-18} \cline{9-18} \cline{10-18} \cline{11-18} \cline{12-18} \cline{13-18} \cline{14-18} \cline{15-18} \cline{16-18} \cline{17-18} \cline{18-18} 
 & EST{[}$\mathcal{R}${]}~\cite{advrank} & 8 & 52.7 & 58.5 & 30.1 & 85.7 & 6.4 & 69.7 & 3.9 & 64.6 & 0.611 & 1.053 & 3.8 & 2.2 & 10.2 & 1.3 & 19.0\tabularnewline
\rowcolor{black!7}\cellcolor{white} & \textcolor{cyan}{ACT{[}$\mathcal{R}${]}}~\cite{robrank} & 8 & 45.3 & 50.6 & 24.1 & 84.7 & 24.8 & 10.7 & 25.4 & 8.2 & 0.321 & 0.485 & 15.4 & 17.7 & 25.1 & 11.3 & 49.5\tabularnewline
& \textcolor{tab:pink}{HM{[}$\mathcal{S},g_{\mathsf{LGA}}${]}} & 8 & 49.0 & 54.1 & 26.4 & 85.0 & 29.9 & 4.7 & 31.6 & 3.6 & 0.455 & 0.283 & 39.3 & 40.9 & 38.8 & 43.0 & 61.7\tabularnewline
\rowcolor{black!7}\cellcolor{white} & \textcolor{crimson}{HM{[}$\mathcal{S},g_{\mathsf{LGA}}${]}\&ICS} & 8 & 48.3 & 53.4 & 25.7 & 84.9 & 32.5 & 4.8 & 32.4 & 3.5 & 0.586 & 0.239 & 38.6 & 39.8 & 38.3 & 44.5 & 61.2\tabularnewline
\cline{2-18} \cline{3-18} \cline{4-18} \cline{5-18} \cline{6-18} \cline{7-18} \cline{8-18} \cline{9-18} \cline{10-18} \cline{11-18} \cline{12-18} \cline{13-18} \cline{14-18} \cline{15-18} \cline{16-18} \cline{17-18} \cline{18-18} 
 & EST{[}$\mathcal{R}${]}~\cite{advrank} & 32 & 46.0 & 51.4 & 24.5 & 84.7 & 12.5 & 43.6 & 10.6 & 34.8 & 0.468 & 0.830 & 9.6 & 7.2 & 17.3 & 3.8 & 31.7\tabularnewline
\rowcolor{black!7}\cellcolor{white} & \textcolor{cyan}{ACT{[}$\mathcal{R}${]}}~\cite{robrank} & 32 & 47.5 & 52.6 & 25.5 & 84.9 & 24.1 & 10.5 & 22.7 & 9.4 & 0.253 & 0.532 & 21.2 & 21.6 & 27.8 & 15.3 & 50.8\tabularnewline
& \textcolor{tab:pink}{HM{[}$\mathcal{S},g_{\mathsf{LGA}}${]}} & 32 & 47.7 & 52.7 & 25.3 & 84.8 & 30.6 & 4.7 & 31.2 & 3.5 & 0.466 & 0.266 & 38.6 & 40.3 & 38.6 & 44.3 & 61.8\tabularnewline
\rowcolor{black!7}\cellcolor{white}\multirow{-9}{*}{SOP} & \textcolor{crimson}{HM{[}$\mathcal{S},g_{\mathsf{LGA}}${]}\&ICS} & 32 & 46.8 & 51.7 & 24.5 & 84.7 & 32.0 & 4.2 & 33.7 & 3.0 & 0.606 & 0.207 & 39.1 & 39.8 & 37.9 & 45.6 & 61.6\tabularnewline

	\bottomrule

\end{tabular}}
	\vspace{-0.8em}
\caption{Comparison of our defense with the state-of-the-art methods on commonly used DML datasets.}
	\label{tab:sota}
	\vspace{-0.4em}
\end{table*}

\subsection{Comparison to State-of-The-Art Defense}
\label{sec:43}

After validating the effectiveness of our proposed method, we conduct
experiments on CUB, CARS and SOP to compare our proposed method
with the state-of-the-art defense methods, \ie, EST~\cite{advrank}
and ACT~\cite{robrank}. The corresponding results are shown in \cref{tab:sota}.
An ideal defense method should be able to achieve a high ERS and a high R@1
at a low training cost (\ie, $\eta+1$).
The ability of a method to achieve a high ERS under a low training cost
indicates a high efficiency.

According to the results, EST[$\mathcal{R}$] achieves a relatively high R@1
when $\eta{=}8$, but suffers from a drastic drop in R@1 when $\eta$ is
increased to $32$.
Nevertheless, EST[$\mathcal{R}$] only lead to a moderate robustness
compared to other methods.
Experiments for EST[$\mathcal{S}$] are omitted as EST has been greatly
outperformed by ACT~\cite{robrank}, and it is expected to result in
even lower ERS based on the observations in previous subsections.
Although ACT[$\mathcal{R}$] achieves a relatively high ERS, its R@1 performance
drop is distinct on every dataset.
According the previous subsections, ACT[$\mathcal{S}$] can lead to a high R@1,
but along with a significantly lower ERS.
Thus, results for ACT[$\mathcal{S}$] are omitted for being insufficiently robust.

Our method overwhelmingly outperforms the previous methods in terms of the
overall performance. Namely, our method efficiently reaches the highest ERS with a very
low decrement in R@1 under a fixed training cost.
HM[$\mathcal{R},\mathcal{M}$] or HM[$\mathcal{R},g_\mathsf{LGA}$] can reach an
even higher ERS, but are excluded from comparison due to significant drop in
R@1.

It \emph{must} be acknowledged that the high R@1 performance of our method
largely stems from the source triplet sampling strategy, \ie, Softhard, instead
of our contribution.
Nevertheless, the state-of-the-art method, \ie, ACT could not reach the same
level of robustness with the same sampling strategy.

It \emph{should} be noted that the $L_\text{ICS}$ term improves 
robustness against most attacks involved in ERS, but also increases the
tendency to collapse (observed during TMA~\cite{flowertower} attack --
high cosine similarity between two arbitrary benign examples).
In some cases (\eg, on SOP), the robustness drop \emph{w.r.t} TMA may
neutralize the ERS gain from other attacks.

Conclusively, being selective on both benign and
adversarial training samples is crucial for preventing model collapse, and
achieving good performance on both types of samples.
HM is a flexible tool for specifying such ``selection'' of adversarial
examples, while LGA can be interpreted as a concrete ``selection''.
ICS loss further exploits the given sextuplet.

Further analysis, technical details, and limitations, \etc,
are presented in the supplementary document.

\section{Conclusion}
\label{sec:5}

%
In this paper, HM efficiently and flexibly creates adversarial examples for
adversarial training;
LGA specifies an ``intermediate'' destination hardness for balancing robustness
and performance on benign examples;
ICS loss term further improves model robustness.
The state-of-the-art defenses have been surpassed in terms of overall performance.


{\small
\bibliographystyle{ieee_fullname}
\bibliography{robdml}
}

\clearpage
\appendix

\setcounter{figure}{8}
\setcounter{table}{6}

\section*{Supplementary Material}

In this supplementary material, we provide more technical details and
additional discussions that are excluded from the manuscript due to space
limit.

\tableofcontents

\section{Additional Information}
\label{sec:a}

\subsection{Potential Societal Impact}
\label{sec:a1}

\noindent\textbf{I. Security.}

Adversarial defenses alleviate the negative societal impact of adversarial
attacks, and hence have positive societal impact.

\subsection{Limitations of Our Method}
\label{sec:a2}

\noindent\textbf{I. Assumptions.}

~\newline
\noindent \ul{(1) Triplet Training Assumption.}

Our method assumes sample triplets are used for training.
Our method may not be compatible to other non-triplet DML loss functions.
Adversarial training with other DML loss functions is left for future study.

~\newline
\noindent \ul{(2) Embedding Space Assumption.}

We follow the common setups~\cite{revisiting,robrank} on the embedding space.
Namely, (1) the embedding vectors are normalized onto the real unit
hypersphere;
(2) the distance function $d(\cdot,\cdot)$ is Euclidean distance.
Our formulations are developed upon the two assumptions.
It is unknown whether our method method will be effective when embedding
vectors are \emph{not} normalized.
And it is unknown whether our method will be effective when $d(\cdot,\cdot)$ is
replaced as other distance metrics, \eg, cosine distance.

~\newline
\noindent \ul{(3) Optimizer Assumption.}

Our method assumes PGD~\cite{madry} is used for optimizing the HM objective
to create adversarial examples.
The Eq.~(4)-(5) may not necessarily hold with other possible optimizers.

~\newline
\noindent\textbf{II. Performance-Sensitive Factors.}

~\newline
\noindent \ul{(1) Maximum number of PGD iterations $\eta$.}

Our method's sensitivity to $\eta$ has been demonstrated by Tab.~3-6 and
Fig.~6-8.
A larger $\eta$ indicates higher training cost, and stronger adversarial
examples are created for adversarial training.
As a result, a larger $\eta$ leads to a higher robustness (ERS) and a lower
R@1 performance.
Our method consistently achieves a higher ERS under different $\eta$ settings
compared to previous methods, and hence are the most efficient defense method.
Experiments with $\eta$ larger than $32$ are not necessary because ERS plateaus
according to Fig.~6-8.

~\newline
\noindent \ul{(2) Source hardness $H_\mathsf{S}$ and destination hardness
$H_\mathsf{D}$.}

The $H_\mathsf{S}$ and $H_\mathsf{D}$ are the only two adjustable items in HM.

The source hardness $H_\mathsf{S}$ depends on triplet sampling strategy.
We conduct experiments with existing triplet sampling strategies in order to
focus on defense.

The choices for $H_\mathsf{D}$ are more flexible than those of $H_\mathsf{S}$,
as discussed in Sec.~3.1.
In the experiments, we study some possible choices following the discussion and
design LGA based on the empirical observations.

~\newline
\noindent\ul{(3) Parameters involved in $g_\mathsf{LGA}$.}

\begin{table*}
\resizebox{\linewidth}{!}{
\setlength{\tabcolsep}{0.47em}
\renewcommand{\arraystretch}{1.1}%
\begin{tabular}{c|cc|cccc|ccccc|ccccc|c}
	\toprule
\multirow{2}{*}{\textbf{Dataset}} & \multirow{2}{*}{\textbf{Defense}} & \multirow{2}{*}{$\eta$} & \multicolumn{4}{c|}{\textbf{Benign Example}} & \multicolumn{10}{c|}{\textbf{White-Box Attacks for Robustness Evaluation}} & \multirow{2}{*}{\textbf{ERS$\uparrow$}}\tabularnewline
\cline{4-17} \cline{5-17} \cline{6-17} \cline{7-17} \cline{8-17} \cline{9-17} \cline{10-17} \cline{11-17} \cline{12-17} \cline{13-17} \cline{14-17} \cline{15-17} \cline{16-17} \cline{17-17}
 &  &  & R@1$\uparrow$ & R@2$\uparrow$ & mAP$\uparrow$ & NMI$\uparrow$ & CA+$\uparrow$ & CA-$\downarrow$ & QA+$\uparrow$ & QA-$\downarrow$ & TMA$\downarrow$ & ES:D$\downarrow$ & ES:R$\uparrow$ & LTM$\uparrow$ & GTM$\uparrow$ & GTT$\uparrow$ & \tabularnewline
 \midrule
\multirow{2}{*}{CUB} & HM{[}$\mathcal{S},g_{\mathsf{LGA}}${]} & 8 & 38.0 & 48.3 & 21.8 & 49.3 & 12.7 & 46.4 & 11.6 & 39.9 & 0.567 & 0.783 & 16.8 & 11.9 & 27.9 & 1.4 & 32.4\tabularnewline
 & HM{[}$\mathcal{S},g_{\mathsf{LGA}}${]} ($u=2.2$) & 8 & 34.8 & 45.5 & 15.2 & 47.1 & 13.4 & 36.0 & 17.2 & 26.1 & 0.934 & 0.244 & 20.1 & 15.9 & 27.3 & 3.8 & 36.1\tabularnewline
 \bottomrule
\end{tabular}}
	\caption{The efficacy of parameter $u$ for clipping loss value $\ell_{t-1}$.
	Stronger adversarial examples will be created for training if the loss value is not
	clipped (equivalent to setting $u$ to the theoretical upper bound of loss,
	\ie, 2.2).
	}
	\label{tab:u}
\end{table*}

A constant $u$ is used to normalize the loss value of the previous training
iteration $\ell_{t-1}$ into $\bar{\ell}_{t-1}\in[0,1]$.
The constant is empirically selected as $u=\gamma$ in our experiment.
According to our observation, the loss value will quickly decrease below
$\gamma$, and will remain in the $[0,\gamma]$ range for the whole training
process.
If we set $u$ to a larger constant than $\gamma$, the normalized loss
$\bar{\ell}_{t-1}$ will be smaller, and results in stronger adversarial
examples through HM[$\mathsf{S},g_\mathsf{LGA}$] and harms the model performance on benign examples,
as shown in \cref{tab:u}.

Another parameter in $g_\mathsf{LGA}$ is the triplet margin parameter $\gamma$
in order to align to the hardness range of Semihard triplets, \ie, $-\gamma<g_\mathsf{LGA}<0$.
We follow the common setup~\cite{revisiting} for this parameter.

~\newline
\noindent\ul{(4) Constant parameter $\lambda$ for ICS loss term $L_\text{ICS}$.}

The weight constant $\lambda$ is set as $0.5$ by default, and $0.05$ on the SOP dataset.
As demonstrated in Tab.~5, there is a trade-off between robustness and
performance on benign examples when tuning the $\lambda$ parameter.

Additional experiments with $\lambda=0.5$ on the SOP dataset can be found in \cref{tab:soplambda}.
Our method is sensitive to this parameter.
An excessively large $\lambda$ on the SOP datasets leads to worse performance on benign examples
and worse robustness.

\begin{table*}
\resizebox{\linewidth}{!}{
\setlength{\tabcolsep}{0.42em}
\renewcommand{\arraystretch}{1.1}%
\begin{tabular}{c|cc|cccc|ccccc|ccccc|c}
	\toprule
\multirow{2}{*}{\textbf{Dataset}} & \multirow{2}{*}{\textbf{Defense}} & \multirow{2}{*}{$\eta$} & \multicolumn{4}{c|}{\textbf{Benign Example}} & \multicolumn{10}{c|}{\textbf{White-Box Attacks for Robustness Evaluation}} & \multirow{2}{*}{\textbf{ERS$\uparrow$}}\tabularnewline
\cline{4-17} \cline{5-17} \cline{6-17} \cline{7-17} \cline{8-17} \cline{9-17} \cline{10-17} \cline{11-17} \cline{12-17} \cline{13-17} \cline{14-17} \cline{15-17} \cline{16-17} \cline{17-17}
 &  &  & R@1$\uparrow$ & R@2$\uparrow$ & mAP$\uparrow$ & NMI$\uparrow$ & CA+$\uparrow$ & CA-$\downarrow$ & QA+$\uparrow$ & QA-$\downarrow$ & TMA$\downarrow$ & ES:D$\downarrow$ & ES:R$\uparrow$ & LTM$\uparrow$ & GTM$\uparrow$ & GTT$\uparrow$ & \tabularnewline
 \midrule
\multirow{2}{*}{SOP} & HM{[}$\mathcal{S},g_{\mathsf{LGA}}${]}\&ICS ($\lambda=0.5$) & 8 & 42.4 & 47.1 & 10.2 & 84.2 & 34.9 & 3.8 & 36.7 & 2.3 & 0.879 & 0.093 & 35.3 & 36.5 & 35.2 & 49.1 & 60.1\tabularnewline
 & HM{[}$\mathcal{S},g_{\mathsf{LGA}}${]}\&ICS ($\lambda=0.5$) & 32 & 41.5 & 46.1 & 9.9 & 84.1 & 36.1 & 3.1 & 37.6 & 2.1 & 0.873 & 0.086 & 35.7 & 36.8 & 34.7 & 50.2 & 60.8\tabularnewline
 \bottomrule
\end{tabular}}
	\caption{An excessively large $\lambda$ may lead to worse performance and robustness.}
	\label{tab:soplambda}
\end{table*}

~\newline
\noindent\ul{(5) Backbone deep neural network.}

We adopt ResNet-18 following the state-of-the-art defense~\cite{robrank} for
fair comparison.
Since our proposed method is independent to the backbone choice, and hence
is expected to be effective with different backbone models.

\subsection{Use of Existing Assets}

\noindent\ul{(1) Datasets.}

All the datasets used in our paper are public datasets,
and their corresponding papers are cited.
The CUB~\cite{cub200} dataset includes images of birds.
The CARS~\cite{cars196} dataset includes images of cars.
The SOP~\cite{sop} dataset includes images of online products.

~\newline
\noindent\ul{(2) Code and Implementation.}

Our implementation is built upon PyTorch
and the public code of the state-of-the-art defense method ACT~\cite{robrank}
(License: Apache-2.0).

\section{Technical Details \& Minor Discussions}
\label{sec:b}

\subsection{Difference between Existing Defenses \& HM}
\label{sec:b1}

\noindent\textbf{I. Embedding-Shifted Triplet (EST).}~\cite{advrank}

Embedding-Shifted Triplet (EST)~\cite{advrank} adopts adversarial counterparts
of $\va,\vp,\vn$ with maximum embedding move distance off their original
locations, \ie,
\begin{equation}
L_\text{EST}=L_\text{T}(\tilde{\va},\tilde{\vp},\tilde{\vn};\gamma)
\end{equation}
where
$\tilde{\va}=\phi(\mA+\vr^*)$, and $\vr^*=\arg\max_{\vr}d_\phi(\mA+\vr, \mA)$.
The $\tilde{\vp}$ and $\tilde{\vn}$ are obtained similarly.

\ul{(1) Relationship with HM:}

Since EST only aims to maximize the embedding move distance off its original
location without specifying any direction, it leads to a random hardness value.
The expectation $E[H(\cdot)]$ of its resulting adversarial triplet is expected
to be close to $E[H_\mathsf{S}]$.
Because the perturbed triplet can be either harder or easier than the benign
triplet.
Namely, EST merely indirectly increase the hardness of the training triplet,
and may even decrease its hardness.
Thus, EST suffers from inefficiency in adversarial training compared to HM.

~\newline
\noindent\textbf{II. Anti-Collapse Triplet (ACT).}~\cite{robrank}

Anti-Collapse Triplet (ACT)~\cite{robrank} ``collapses'' the embedding vectors
of positive and negative sample, and enforces the model to separate them apart,
\ie,
\begin{align}
	L_\text{ACT} &=L_\text{T}(\va, \overrightarrow{\vp},
\overleftarrow{\vn};\gamma),\\
	[\overrightarrow{\vp},\overleftarrow{\vn}] & =[\phi(\mP+\vr_p^*), \phi(\mN+\vr_n^*)]\\
	[\vr_p^*,\vr_n^*] &= \argmin_{\vr_p,\vr_n} d_\phi(\mP+\vr_p, \mN+\vr_n).
\end{align}

\ul{(1) Relationship with HM:}

When ACT successively ``collapses'' the positive and negative embedding
vectors together, the hardness will be zero, \ie, $E[H(\cdot)]=0$.
But ACT is not equivalent to $HM[\cdot,0]$ because the two methods have
different objectives and use different gradients.
Besides, in order to avoid the ``misleading gradients''~\cite{robrank}, ACT
fixes the anchor and only perturb the positive and negative samples, which
makes the objective for creating adversarial examples more difficult to
optimize in practice.
In brief, ACT is also indirectly increasing the loss value, suffering from
inefficient adversarial learning.

\begin{table*}[t]
\resizebox{\linewidth}{!}{
\setlength{\tabcolsep}{0.47em}
\renewcommand{\arraystretch}{1.1}%
\begin{tabular}{c|cc|cccc|ccccc|ccccc|c}
	\toprule
\multirow{2}{*}{\textbf{Dataset}} & \multirow{2}{*}{\textbf{Defense}} & \multirow{2}{*}{$\eta$} & \multicolumn{4}{c|}{\textbf{Benign Example}} & \multicolumn{10}{c|}{\textbf{White-Box Attacks for Robustness Evaluation}} & \multirow{2}{*}{\textbf{ERS$\uparrow$}}\tabularnewline
\cline{4-17} \cline{5-17} \cline{6-17} \cline{7-17} \cline{8-17} \cline{9-17} \cline{10-17} \cline{11-17} \cline{12-17} \cline{13-17} \cline{14-17} \cline{15-17} \cline{16-17} \cline{17-17}
 &  &  & R@1$\uparrow$ & R@2$\uparrow$ & mAP$\uparrow$ & NMI$\uparrow$ & CA+$\uparrow$ & CA-$\downarrow$ & QA+$\uparrow$ & QA-$\downarrow$ & TMA$\downarrow$ & ES:D$\downarrow$ & ES:R$\uparrow$ & LTM$\uparrow$ & GTM$\uparrow$ & GTT$\uparrow$ & \tabularnewline
 \midrule
\multirow{3}{*}{CUB} & HM{[}$\mathcal{S},g_{\mathsf{1/2}}${]} & 8 & 38.7 & 48.5 & 22.0 & 49.2 & 12.6 & 48.6 & 12.3 & 41.3 & 0.562 & 0.825 & 13.5 & 12.9 & 26.9 & 1.8 & 31.7\tabularnewline
& HM{[}$\mathcal{S},g_{\mathsf{LGA}}${]} & 8 & 38.0 & 48.3 & 21.8 & 49.3 & 12.7 & 46.4 & 11.6 & 39.9 & 0.567 & 0.783 & 16.8 & 11.9 & 27.9 & 1.4 & 32.4\tabularnewline
& HM{[}$\mathcal{S},g_{\mathsf{2}}${]} & 8 & 37.4 & 48.2 & 17.1 & 49.2 & 12.9 & 45.1 & 13.2 & 39.1 & 0.599 & 0.738 & 17.7 & 12.8 & 27.5 & 1.9 & 33.1\tabularnewline
\bottomrule
\end{tabular}}
	\caption{Non-linear Gradual Adversary Examples.}
	\label{tab:nonlinga}
\end{table*}

~\newline\textbf{III. Min-max Adversarial Training with Triplet Loss.}

The direct formulation of min-max adversarial training~\cite{madry} for
triplet loss-based DML is:
\begin{equation}
	\theta^* = \argmin_\theta [ \argmax_{\vr_a, \vr_p, \vr_n} L_\text{T}(
	\mA + \vr_a, \mP + \vr_p, \mN + \vr_n)]
\end{equation}
Previous works~\cite{advrank,robrank} point out this method will easily lead
to model collapse.
Our observation suggests the same.

\ul{(1) Relationship with HM:}

Maximizing $L_\text{T}$ is equivalent to maximizing $\tilde{H}_\mathsf{S}$.
This can be expressed as HM[$H_\mathsf{S},2$] as discussed in Sec.~3.1.

\subsection{Hardness Manipulation (HM)}
\label{sec:b2}

\noindent\textbf{I. Adjustable Items}

The only two adjustable items in HM are $H_\mathsf{S}$ and $H_\mathsf{D}$.
They are discussed in the ``Performance-Sensitive Factors'' part of the
previous section.

~\newline
\noindent\textbf{II. Extreme Values of Hardness.}

As reflected in Sec.~3.1 and Fig.~2, the range of $H(\cdot)$ is $[-2, 2]$.
The embedding vectors have been normalized to the real unit hypersphere
as pointed out in the manuscript.
And the range of distance between any two points on the hypersphere is
$[0,2]$.
Hence the extreme values are:
\begin{align}
	& \max H(\mA, \mP, \mN)\\
	=& \max [d_\phi(\mA, \mP) - d_\phi(\mA, \mN)]\\
	=& \max [d_\phi(\mA, \mP)] - \min [d_\phi(\mA, \mN)]\\
	=& 2 - 0,
\end{align}
\begin{align}
	& \min H(\mA, \mP, \mN)\\
	=& \min [d_\phi(\mA, \mP) - d_\phi(\mA, \mN)]\\
	=& \min [d_\phi(\mA, \mP)] - \max [d_\phi(\mA, \mN)]\\
	=& 0 - 2.
\end{align}
Namely $H(\cdot)\in[-2,2]$. Meanwhile, since
\begin{equation}
L_\text{T}(\mA, \mP, \mN; \gamma) = \max(0, H(\mA, \mP, \mN) + \gamma),
\end{equation}
we have $L_\text{T}\in[0, 2+\gamma]$.

\subsection{Gradual Adversary}
\label{sec:b3}

\noindent\textbf{I. Parameters}

The parameters, namely $u$ and $\gamma$ are discussed in the previous section
of this supplementary material, see ``Performance-Sensitive Factors''.

The parameter $\xi$ in $g_\mathsf{B}$ is set as $0.1$, but it is not an important parameter.
The function $g_\mathsf{B}$ is only used for demonstrating that ``slightly
boosting the destination hardness can further increase ERS'' as discussed
in Sec.~3.1.

~\newline
\noindent\textbf{II. Non-linear Gradual Adversary}

In Sec.~3.2, more complicated designs are left for future work.
We provide two Non-linear Gradual Adversary examples, namely
$g_\mathsf{2}$ and $g_\mathsf{1/2}$, as follows:
\begin{align}
	g_\mathsf{2}(\cdot)   &= -\gamma \cdot (\bar{\ell}_{t-1})^{2}  &~ \in [-\gamma,0]\\
	g_\mathsf{1/2}(\cdot) &= -\gamma \cdot (\bar{\ell}_{t-1})^{1/2} &~ \in [-\gamma, 0]
\end{align}
Compared to LGA, $g_\mathsf{2}$ is more ``eager'' to result in strong adversarial
examples in the early phase of training, while $g_\mathsf{1/2}$ is more ``conservative''
in creating strong adversarial examples in the early phase of training
(adversarial examples from HM are stronger if the function value is closer to $0$).
The corresponding experiments can be found in \cref{tab:nonlinga}.

\begin{table*}
\resizebox{\linewidth}{!}{
\setlength{\tabcolsep}{0.47em}
\renewcommand{\arraystretch}{1.1}%
\begin{tabular}{c|cc|cccc|ccccc|ccccc|c}
	\toprule
\multicolumn{1}{c|}{\multirow{2}{*}{\textbf{Dataset}}} & \multirow{2}{*}{\textbf{Defense}} & \multirow{2}{*}{$\eta$} & \multicolumn{4}{c|}{\textbf{Benign Example}} & \multicolumn{10}{c|}{\textbf{White-Box Attacks for Robustness Evaluation}} & \multirow{2}{*}{\textbf{ERS$\uparrow$}}\tabularnewline
\cline{4-17} \cline{5-17} \cline{6-17} \cline{7-17} \cline{8-17} \cline{9-17} \cline{10-17} \cline{11-17} \cline{12-17} \cline{13-17} \cline{14-17} \cline{15-17} \cline{16-17} \cline{17-17}
 &  &  & R@1$\uparrow$ & R@2$\uparrow$ & mAP$\uparrow$ & NMI$\uparrow$ & CA+$\uparrow$ & CA-$\downarrow$ & QA+$\uparrow$ & QA-$\downarrow$ & TMA$\downarrow$ & ES:D$\downarrow$ & ES:R$\uparrow$ & LTM$\uparrow$ & GTM$\uparrow$ & GTT$\uparrow$ & \tabularnewline
 \midrule
\multirow{2}{*}{CUB} & HM{[}$\mathcal{S},g_{\mathsf{LGA}}${]}\&ICS & 8 & 37.2 & 47.8 & 21.4 & 48.4 & 12.9 & 40.9 & 14.7 & 33.7 & 0.806 & 0.487 & 17.1 & 13.2 & 26.3 & 2.3 & 33.5\tabularnewline
& HM[$\mathcal{S},g_\mathsf{LGA}$]\&$L_\text{ICS}(\cdots;\gamma=0.2)$ & 8 & 35.8 & 46.6 & 16.4 & 48.3 & 13.2 & 39.9 & 12.7 & 33.3 & 0.775 & 0.507 & 15.9 & 14.9 & 27.2 & 2.7 & 33.6\tabularnewline
 \bottomrule
\end{tabular}}
	\caption{The margin parameter is set to $0$ in order to avoid negative effect.
	R@1 performance drops with marginal robustness gain.}
	\label{tab:0margin}
\end{table*}

\subsection{Intra-Class Structure (ICS)}

\noindent\textbf{I. Parameters}

The ICS loss term can be appended to the loss for adversarial training.
The only parameter for ICS loss term is the weight constant $\lambda$,
and has been discussed in the ``Performance-Sensitive Factors'' part
of the previous section.

The margin parameter is set to $0$ in order to avoid negative effect in $L_\text{ICS}$,
as shown in \cref{tab:0margin}.

~\newline
\noindent\textbf{II. Gradients in Fig.~5 (a) in Manuscript}

According to \cite{robrank}, when the embedding vectors are normalized onto the
real unit hypersphere and Euclidean distance is used, the gradients of the
triplet loss with respect to the anchor, positive, and negative embedding
vectors are respectively:
\begin{align}
	\frac{\partial L_\text{T}}{\partial \va} &= \frac{\va - \vp}{\|\va - \vp\|}
	- \frac{\va - \vn}{\| \va - \vn \|} \\
	\frac{\partial L_\text{T}}{\partial \vp} &= \frac{\vp - \va}{\|\va - \vp\|} \\
	\frac{\partial L_\text{T}}{\partial \vn} &= \frac{\va - \vn}{\|\va - \vn\|},
\end{align}
when $L_\text{T}>0$.
And the above equations have been reflected in Fig.~5 (a) in terms of vector direction.

~\newline
\noindent\textbf{III. Alternative Design for Exploiting Sextuplet}

Let $\tilde{\va}$, $\tilde{\vp}$, and $\tilde{\vn}$ be $\phi(\mA+\hat{\vr}_a)$,
$\phi(\mP+\hat{\vr}_p)$ and $\phi(\mN+\hat{\vr}_n)$ respectively.
In our proposed ICS loss term, only $(\va, \tilde{\va}, \vp)$ are involved.
Other alternative selections of triplets from the sextuplet are possible,
but are not as effective as $L_\text{ICS}$ for improving adversarial robustness.

\ul{(1) $L_\text{T}(\va, \tilde{\va}, \tilde{\vp})$}

As shown in Fig.~5 (c) in the manuscript, the position of $\tilde{\vp}$ is
always further away from both $\va$ and $\tilde{\va}$ than $\vp$ due to
the gradient direction.
Thus, the loss value of $L_\text{T}(\va, \tilde{\va}, \tilde{\vp})$ will always
be smaller than $L_\text{ICS}$, and hence is less effective than $L_\text{ICS}$.

\ul{(2) $L_\text{T}(\va, \vp, \vn)$: Mixing regular training
and adversarial training.}

According to our observation, mixing regular training and adversarial training
leads to better R@1 with drastic robustness degradation for both ACT and
our defense.

\ul{(3) $L_\text{T}(\vp, \tilde{\vp}, \va)$: Symmetric counterpart of $L_\text{ICS}$.}

Every sample in the training dataset will be used as anchor for once per
epoch.
Such symmetric loss term is duplicated to $L_\text{ICS}$ and is
not necessary.
Experimental results suggest negligible difference compared to $L_\text{ICS}$.

\ul{(4) $L_\text{T}(\va, \tilde{\va}, \tilde{\vn})$: $\tilde{\vn}$ is very close to $\va$ in Fig.~5.}

It enforces \emph{inter}-class structure instead of intra-class structure.
Besides, experimental results suggest negligible difference.
We speculate this loss term is duplicated to the adversarial training loss term,
\ie, $L_\text{T}(\tilde{\va}, \tilde{\vp}, \tilde{\vn})$, which enforces
inter-class structure as well in a stronger manner.

\ul{(5) $L_\text{T}(\va, \vp, \tilde{\vn})$}

According to our observation, it leads to better R@1 performance on benign
examples, but drastically reduce the robustness.

\subsection{Experiments and Evaluation}

\noindent\textbf{I Hardware and Software Configuration}

We conduct the experiments with two Nvidia Titan Xp GPUs (12GB of memory each)
under the Ubuntu 16.04 operating system with PyTorch 1.8.2 in distributed data
parallel.

~\newline
\noindent\textbf{II. Training Cost}

In the manuscript, the training cost of adversarial training is caluclated as
$\eta+1$, which is the number of forward-backward propagation involved in each
iteration of the training process, in order to reflect the training efficiency
(gain as high robustness as possible given a fixed number of forward-backward
calculation for adversarial training) of different defense methods.
Specifically, PGD creates adversarial examples from the $\eta$ times of
forward-backward computation.
With the resulting adversarial examples, the network requires once
more forward-backward computation to update the model parameters.

~\newline
\noindent\textbf{III. Complete Results for Tab.~2 in Manuscript}

\begin{table*}
\resizebox{\linewidth}{!}{
\setlength{\tabcolsep}{0.48em}%
\begin{tabular}{c|cc|cccc|ccccc|ccccc|c}

	\toprule

\multirow{2}{*}{\textbf{Dataset}} & \multirow{2}{*}{\textbf{Defense}} & \multirow{2}{*}{$\eta$} & \multicolumn{4}{c|}{\textbf{Benign Example}} & \multicolumn{10}{c|}{\textbf{White-Box Attacks for Robustness Evaluation}} & \multirow{2}{*}{\textbf{ERS}}\tabularnewline
\cline{4-17} \cline{5-17} \cline{6-17} \cline{7-17} \cline{8-17} \cline{9-17} \cline{10-17} \cline{11-17} \cline{12-17} \cline{13-17} \cline{14-17} \cline{15-17} \cline{16-17} \cline{17-17} 
 &  &  & R@1$\uparrow$ & R@2$\uparrow$ & mAP$\uparrow$ & NMI$\uparrow$ & CA+$\uparrow$ & CA-$\downarrow$ & QA+$\uparrow$ & QA-$\downarrow$ & TMA$\downarrow$ & ES:D$\downarrow$ & ES:R$\uparrow$ & LTM$\uparrow$ & GTM$\uparrow$ & GTT$\uparrow$ & \tabularnewline

	\midrule

\multirow{5}{*}{CUB} & HM{[}$\mathcal{R},\mathcal{R}${]} & N/A & 53.9 & 66.4 & 26.1 & 59.5 & 0.0 & 100.0 & 0.0 & 99.9 & 0.883 & 1.762 & 0.0 & 0.0 & 14.1 & 0.0 & 3.8\tabularnewline
 & HM{[}$\mathcal{R},\mathcal{M}${]} & 8 & 27.0 & 36.0 & 13.2 & 42.5 & 19.4 & 48.0 & 22.2 & 32.0 & 0.535 & 0.867 & 11.6 & 10.4 & 19.3 & 2.9 & 35.1\tabularnewline
 & HM{[}$\mathcal{R},\mathcal{S}${]} & 8 & C & o & l & l & a & p & s & e &  &  &  &  &  &  & \tabularnewline
 & HM{[}$\mathcal{R},\mathcal{D}${]} & 8 & C & o & l & l & a & p & s & e &  &  &  &  &  &  & \tabularnewline
 & HM{[}$\mathcal{R},\mathcal{H}${]} & 8 & C & o & l & l & a & p & s & e &  &  &  &  &  &  & \tabularnewline
\hline 
\multirow{5}{*}{CUB} & HM{[}$\mathcal{M},\mathcal{R}${]} & 8 & 43.9 & 55.9 & 27.1 & 54.2 & 0.2 & 100.0 & 0.4 & 99.6 & 0.849 & 1.640 & 0.2 & 0.1 & 19.1 & 0.0 & 5.4\tabularnewline
 & HM{[}$\mathcal{M},\mathcal{M}${]} & N/A & 44.0 & 56.1 & 27.2 & 54.6 & 0.1 & 100.0 & 0.4 & 99.8 & 0.843 & 1.717 & 0.3 & 0.3 & 18.4 & 0.0 & 5.0\tabularnewline
 & HM{[}$\mathcal{M},\mathcal{S}${]} & 8 & C & o & l & l & a & p & s & e &  &  &  &  &  &  & \tabularnewline
 & HM{[}$\mathcal{M},\mathcal{D}${]} & 8 & C & o & l & l & a & p & s & e &  &  &  &  &  &  & \tabularnewline
 & HM{[}$\mathcal{M},\mathcal{H}${]} & 8 & C & o & l & l & a & p & s & e &  &  &  &  &  &  & \tabularnewline
\hline 
\multirow{5}{*}{CUB} & HM{[}$\mathcal{S},\mathcal{R}${]} & 8 & 48.3 & 60.1 & 30.3 & 56.3 & 2.0 & 84.4 & 1.4 & 85.4 & 0.835 & 0.895 & 6.1 & 1.6 & 21.0 & 0.0 & 13.7\tabularnewline
 & HM{[}$\mathcal{S},\mathcal{M}${]} & 8 & 38.4 & 49.7 & 22.9 & 50.3 & 10.9 & 50.5 & 10.8 & 44.6 & 0.680 & 0.722 & 13.3 & 11.2 & 25.8 & 1.2 & 29.6\tabularnewline
 & HM{[}$\mathcal{S},\mathcal{S}${]} & N/A & 55.7 & 68.2 & 35.8 & 61.0 & 0.0 & 100.0 & 0.0 & 99.6 & 0.942 & 1.272 & 0.0 & 0.0 & 19.2 & 0.0 & 6.2\tabularnewline
 & HM{[}$\mathcal{S},\mathcal{D}${]} & 8 & C & o & l & l & a & p & s & e &  &  &  &  &  &  & \tabularnewline
 & HM{[}$\mathcal{S},\mathcal{H}${]} & 8 & C & o & l & l & a & p & s & e &  &  &  &  &  &  & \tabularnewline
\hline 
\multirow{5}{*}{CUB} & HM{[}$\mathcal{D},\mathcal{R}${]} & 8 & 52.7 & 64.0 & 33.7 & 58.9 & 0.0 & 100.0 & 0.2 & 100.0 & 0.841 & 1.749 & 0.1 & 0.0 & 19.0 & 0.0 & 4.8\tabularnewline
 & HM{[}$\mathcal{D},\mathcal{M}${]} & 8 & 50.7 & 63.8 & 33.1 & 58.7 & 0.0 & 100.0 & 0.3 & 99.9 & 0.841 & 1.744 & 0.2 & 0.0 & 18.5 & 0.0 & 4.8\tabularnewline
 & HM{[}$\mathcal{D},\mathcal{S}${]} & 8 & C & o & l & l & a & p & s & e &  &  &  &  &  &  & \tabularnewline
 & HM{[}$\mathcal{D},\mathcal{D}${]} & N/A & 51.4 & 64.2 & 33.4 & 59.3 & 0.0 & 100.0 & 0.4 & 99.9 & 0.842 & 1.753 & 0.0 & 0.0 & 19.7 & 0.0 & 4.9\tabularnewline
 & HM{[}$\mathcal{D},\mathcal{H}${]} & 8 & 54.7 & 66.9 & 36.2 & 60.0 & 0.0 & 100.0 & 0.2 & 99.8 & 0.874 & 1.569 & 0.4 & 0.0 & 18.6 & 0.0 & 5.4\tabularnewline
\hline 
\multirow{5}{*}{CUB} & HM{[}$\mathcal{H},\mathcal{R}${]} & 8 & 51.0 & 63.5 & 32.9 & 59.2 & 0.0 & 100.0 & 0.2 & 99.9 & 0.837 & 1.750 & 0.0 & 0.0 & 17.7 & 0.0 & 4.7\tabularnewline
 & HM{[}$\mathcal{H},\mathcal{M}${]} & 8 & 52.2 & 64.0 & 33.1 & 58.4 & 0.0 & 100.0 & 0.2 & 100.0 & 0.841 & 1.750 & 0.0 & 0.0 & 18.9 & 0.0 & 4.8\tabularnewline
 & HM{[}$\mathcal{H},\mathcal{S}${]} & 8 & C & o & l & l & a & p & s & e &  &  &  &  &  &  & \tabularnewline
 & HM{[}$\mathcal{H},\mathcal{D}${]} & 8 & 52.6 & 64.9 & 34.4 & 60.0 & 0.0 & 100.0 & 0.2 & 99.9 & 0.835 & 1.720 & 0.1 & 0.0 & 20.0 & 0.0 & 5.1\tabularnewline
 & HM{[}$\mathcal{H},\mathcal{H}${]} & N/A & 48.9 & 61.2 & 30.3 & 56.5 & 0.0 & 100.0 & 0.4 & 99.9 & 0.830 & 1.752 & 0.2 & 0.0 & 19.5 & 0.0 & 5.0\tabularnewline

	\bottomrule

\end{tabular}}

	\caption{Full data sheet for Tab.2 in the Paper. The symbols $\mathcal{R},
	\mathcal{M}, \mathcal{S}, \mathcal{D}, \mathcal{H}$ denote Random, Semihard,
	Softhard, Distance and Hardest triplet sampling strategies, respectively.
	Experiments that end up with model collapse are left with a placeholder 
	in the table.
	}
\label{tab:srcdest}
\end{table*}

Complete experimental results for ``Tab.~2: Combinations of Source \&
Destination Hardness. \ldots'' can be found in \cref{tab:srcdest} of this
supplementary material.

~\newline
\noindent\textbf{IV. Additional Notes on the Experimental Results}

\noindent\ul{(1) Slight ERS decrease with a larger $\eta$}

In some cases, \eg, HM[$\mathcal{S},g_\mathsf{LGA}$]\&ICS on the CARS dataset
reaches a slightly lower ERS with $\eta=32$ compared to that with $\eta=8$.
We speculate this is because adversarial training suffers from overfitting~\cite{bagoftricks,overfitting}
on adversarial examples, as mentioned in Sec.~2.

\subsection{Potential Future Work}

\noindent\textbf{I. Faster Adversarial Training with HM}

As discussed in Sec.~3.1, one potential adversarial training acceleration
method is to incorporate Free Adversarial Training (FAT)~\cite{freeat}
(originally for classification) into our DML adversarial training with HM.
Besides, directly incorporating FAT into the min-max adversarial training of
DML will easily result in model collapse as well, because the FAT algorithm can
be interpreted as to maintain a universal (agnostic to sample) perturbation
that can maximize the loss.
Thus, non-trivial modifications are still required to incorporate FAT FAT into
adversarial training with HM. This is left for future work.

~\newline\textbf{II. Better Choice for Destination Hardness}

The proposed LGA function incorporates our empirical observation that
``adversarial triplets should remain Semihard'' based on the results in Tab.~2.
However, a better choice for $H_\mathsf{D}$ may exist between ``Semihard'' and
``Softhard'' that can achieve better overall performance.
In the manuscript, we only use the existing sampling methods and simple
pseudo-hardness functions in order to avoid distraction from our focus.

~\newline\textbf{III. Other Loss Functions}

Adversarial trainig with DML loss functions other than triplet loss is
insufficiently explored.
New metric learning loss functions oriented for adversarial training are also
left for future study.
Besides, model collapse is an inevitable problem for adversarial training
with triplet loss, and it is unknown whether other loss functions could
mitigate this issue.

~\newline\textbf{IV. DML \& Classification}

It is unknown whether DML defenses will improve the robustness in the
classification task.

\end{document}